%% file: main.tex
\documentclass[letterpaper,twocolumn,10pt]{article}
\usepackage[left=0.7in, right=0.7in, bottom=0.7in, top=0.7in]{geometry}
\usepackage{mathptmx} 

\usepackage{fancyhdr}
\usepackage[normalem]{ulem}
\usepackage[hyphens]{url}
\usepackage[sort,nocompress]{cite}
\usepackage[final]{microtype}
\usepackage{flushend}
\usepackage{hyperref}
\usepackage[hyphenbreaks]{breakurl}

\usepackage{pdflscape}
\usepackage{longtable}
\usepackage[export]{adjustbox}
\usepackage{subfig}
\usepackage{graphicx}
\usepackage{auto-pst-pdf}
\usepackage{enumerate}
\usepackage{booktabs}
\usepackage{multirow}
\usepackage{array}
\newcolumntype{L}[1]{>{\raggedright\let\newline\\\arraybackslash\hspace{0pt}}m{#1}}
\newcolumntype{C}[1]{>{\centering\let\newline\\\arraybackslash\hspace{0pt}}m{#1}}
\newcolumntype{R}[1]{>{\raggedleft\let\newline\\\arraybackslash\hspace{0pt}}m{#1}}
\usepackage{xcolor,colortbl}
\definecolor{LightBlue}{rgb}{0.83, 0.91, 1}

\fancypagestyle{firstpage}{
  \fancyhf{}
  
  \fancyfoot[C]{\thepage}
}

\title{Demystifying the MLPerf Benchmark Suite}
\author{
{
\rm 
Snehil Verma$^1$,
Qinzhe Wu$^1$,
Bagus Hanindhito$^1$,
Gunjan Jha$^2$
}\\
{
Eugene B. John$^2$,
Ramesh Radhakrishnan$^3$,
and
Lizy K. John$^1$
}\\
University of Texas at Austin$^1$ 
\\University of Texas at San Antonio$^2$
\\Dell Inc.$^3$
} 

\begin{document}
\maketitle
\thispagestyle{firstpage}
\pagestyle{plain}

\begin{abstract}
MLPerf, an emerging machine learning benchmark suite strives to cover a broad range of applications of machine learning.  We present a study on its characteristics and how the MLPerf benchmarks differ from some of the previous deep learning benchmarks like DAWNBench and DeepBench. 
We find that application benchmarks such as MLPerf (although rich in kernels) exhibit different features compared to kernel benchmarks such as DeepBench.
MLPerf benchmark suite contains a diverse set of models which allows unveiling various bottlenecks in the system. Based on our findings, dedicated low latency interconnect between GPUs in multi-GPU systems is required for optimal distributed deep learning training.
We also observe variation in scaling efficiency across the MLPerf models.
The variation exhibited by the different models highlight the importance of smart scheduling strategies for multi-GPU training. 
Another observation is that CPU utilization increases with increase in number of GPUs used for training.
Corroborating prior work we also observe and quantify improvements possible by compiler optimizations, mixed-precision training and use of Tensor Cores.

\end{abstract}

\section{Introduction}\label{sec:introduction}
\input{introduction}

\section{Background}\label{sec:background}
\input{background}

\section{Methodology}\label{sec:methodology}
\input{methodology}

\section{Benchmark Analysis}\label{sec:analysis}
\input{analysis}
\section{System Level Measurements}\label{sec:sys_measurements}
\input{sys_measurements}

\section{Conclusion}\label{sec:conclusion}
\input{conclusion}



\bibliographystyle{IEEEtranS}
\bibliography{refs}

\newpage
\section*{Appendix}
\input{classification}


\end{document}

%% file: introduction.tex
The recent advances in machine learning have led to an evolution of a myriad of applications, revolutionizing scientific, industrial and commercial fields. Machine learning, primarily deep learning, is the state-of-the-art in providing models, methods, tools, and techniques for developing autonomous and intelligent systems. 

There are two parts to machine learning: training and inference. Training refers to the process where the neural network learns a new capability based on existing data. Training is a compute-intensive task since it operates typically on massive datasets, tuning weights until the model meets the desired quality. As the system's compute power plays a significant role in accelerating the neural network learning, training is usually done using high-performance hardware or compute clusters. However, reducing the training time while maintaining the desired quality of the neural network model is still an active area of research. On the other hand, inference utilizes the capabilities of a trained neural network to make useful predictions which requires less compute power. Inference is usually performed inside the end-user hardware such as edge devices where energy efficiency is also an important design consideration.

Benchmarking various machine learning workloads and evaluating their performance based on a reasonable metric is a prerequisite for a fair comparison. MLPerf is an emerging consortium that provides an extensive benchmark suite for measuring the performance of machine learning software frameworks, hardware accelerators, and cloud platforms ~\cite{MLPerf, cliff, microproc-report}. The major contributors include Google, NVIDIA, Baidu, Intel, AMD, and other commercial vendors, as well as research universities such as Harvard, Stanford, and the University of California, Berkeley. MLPerf initial release \texttt{v0.5} consists of benchmarks only for training, with inference benchmarks expected shortly. The benchmarks provide reference implementations for workloads in the areas of vision, language, product recommendation and other key areas where Deep Learning models have shown success and datasets are available publicly. 


Young~\cite{cliff} rightly points out five main attributes that a good machine learning benchmark suite should possess, grouped together as five "R"s. One of them was \textit{Representative workloads}, with regards to which he wrote:

\begin{quote}
    A good benchmark suite is both diverse and representative, where each workload in the suite has unique attributes and the suite collectively covers a large fraction of the application space.
\end{quote}
In a recent talk on "MLPerf design challenges", Mattson~\cite{MLPerf_design} highlighted that the current set of training benchmarks cover a wide range of applications.

\begin{table*}[htb]
\renewcommand{\arraystretch}{1.1}
\centering
\caption{Summary of key insights from the work.} 
\label{table:contributions}
\vspace{-3mm}
\begin{tabular}{|L{0.415\textwidth}|L{0.075\textwidth}|L{0.44\textwidth}|}
\hline
\rowcolor{LightBlue}
\textbf{Observation} & \textbf{Location} & \textbf{Insight/Explanation} \\ \hline
MLPerf benchmark suite has a disjoint envelope from DAWNBench and DeepBench.  & 
Figure~\ref{fig:PC1PC2} &
\multirow{2}{0.44\textwidth}[1mm]{MLPerf, DAWNBench, and DeepBench suite stress HBM2 memory at different levels, and are optimized to different extents.
Throughput and arithmetic intensity: DAWNBench $>$ MLPerf $>$ DeepBench.} \\ 
\cline{1-2}
DeepBench, MLPerf, and DAWNBench are located in different regions in the roofline graph. &
Figure~\ref{fig:roofline} & \\ \hline
Every benchmark in MLPerf benchmark suite is on the boundary of the workload space. &
Figure~\ref{fig:PC3PC4} &
\multirow{2}{0.44\textwidth}[1mm]{There is a great diversity existing in MLPerf benchmark suite, e.g., in terms of the scaling efficiency. This information is helpful for resource scheduling in systems with multiple devices, such as data centers and cloud platforms.}
\\ \cline{1-2}
Different benchmarks scale up differently, and by exploiting these differences, the optimal scheduling can save hours of training on multi-GPU systems.&
Table~\ref{table:Scalability} Figure~\ref{fig:schedules} &
\\ \hline
The linkage distance between Res50\_\textbf{MX} and Res50\_\textbf{TF} is as much as the longest distance among a group of other workloads on different platforms, different application domains (including Res18\_\textbf{Py}) &
Figure~\ref{fig:dendrogram} &
Same neural network model does not guarantee that the characteristics of the workloads would be similar. Hyperparameters and the implementation frameworks largely affect the behavior of the benchmark.
\\ \hline
Data points representing machine learning workloads are close to the slanted roof line. & Figure~\ref{fig:roofline} &
It's easy to exploit the abundant parallelism in ML applications and finally end up being bound by hardware resources
\\ \hline
Mixed precision in combination with TensorCores earns significant speedup on MLPerf. &
Figure~\ref{fig:precision} & 
Hardware support for reduced precision arithmetics is important, especially for machine learning workloads.
\\ \hline
With XLA enabled, Res50\_TF converges to the same accuracy as no XLA, while the time reduces by 40\% &
Figure~\ref{fig:Res50_TTA_frameworks} &
Compiler optimizations, especially kernel fusion, provides for a lot of potential for performance improvement. \\ \hline
When scaling to more GPUs, many benchmarks have a super-linear increase in PCIe / NVLink utilization. & Table~\ref{table:utilization} &
\multirow{2}{0.44\textwidth}[-1mm]{Machine learning applications can become communication-heavy workloads, so the bus is worth attention in ML processor designs. Direct connections between GPUs facilitate machine learning workloads.}
\\
\cline{1-2}
Training time: GPU-system with NVLink enabled $<$ GPU-system with PCIe switch enabled $<$ system with GPUs connected using CPU PCIe ports. &
Figure~\ref{fig:Topology_comp} Table~\ref{table:hardware} & \\ \hline
\end{tabular}
\vspace{-6mm}
\end{table*}

We evaluate the MLPerf benchmarks with experiments on diverse hardware platforms. We investigate whether the execution characteristics of these benchmarks also point out sufficient dissimilarities or are they largely similar in spite of diverse domains?
This work focuses solely on the training workloads. The objective of this work is to unfold the answers to following enigmas:
\begin{itemize}
\setlength\itemsep{.1em}
    \item How different are the MLPerf benchmarks from the prior deep learning benchmarks? How different are the MLPerf benchmarks from each other?
    If hardware designers do not have time budget to evaluate all benchmarks, can they use a subset of the benchmarks?
    \item What is the performance differential that can be obtained by using reduced precision and NVIDIA's tensor cores?
    \item How well does the training performance scale with increasing the number of GPUs? 
    \item What is an efficient way for a user to operate on multiple GPUs to train several models: should they run distributed jobs one-by-one on all GPUs or should they run jobs assigning one model to each GPU or is there any other better solution?
    
    \item How are CPU, GPU and interconnect utilizations? 
     Is there a significant performance impact from the high-bandwidth GPU interconnects?
\end{itemize}

\input{mlperf_summary_table}

{The key insights revealed by this work are summarized in Table~\ref{table:contributions}}, and the rest of paper is organized as follows: Section~\ref{sec:background} introduces the emerging MLPerf~\cite{MLPerf} benchmark suite as well as some prior deep learning benchmarks like DAWNBench~\cite{Coleman2017DAWNBenchA} and DeepBench~\cite{DeepBench}.
In Section~\ref{sec:methodology}, we expand on the system configurations and topologies, on which various experiments were performed. We investigate various benchmark characteristics and present a detailed  analysis of the similarity of various benchmarks in Section~\ref{sec:analysis}. This section also presents performance impact of mixed precision and tensor cores. Section~\ref{sec:sys_measurements} presents measurements on the system resources utilization to provide insights on CPU's, GPU's, and interconnect's impact on machine learning training performance as well as the memory requirement to store the dataset during processing. 
Then, we conclude the paper with Section~\ref{sec:conclusion}.

%% file: mlperf_summary_table.tex

\begin{table*}[t]
\renewcommand{\arraystretch}{1.2}
\centering
\caption{Summary of benchmarks in MLPerf \texttt{v0.5}, DAWNBench, and DeepBench suites used in this study.}
\label{my-label}
\resizebox{\textwidth}{!}{%
\begin{tabular}{|c|c|c|c|c|c|}
\hline
\rowcolor{LightBlue} 
\textbf{Abbreviation} & \textbf{Domain} & \textbf{Model} & \textbf{Framework (submitter)} &  \textbf{Dataset} & \textbf{Quality Target} \\
\hline
\multicolumn{6}{|c|}{\cellcolor[HTML]{ECF4FF}{\textbf{MLPerf} \texttt{v0.5}}} \\ 
\hline
MLPf\_Res50\_TF & &  & TensorFlow (Google) &  &  \\
\cline{1-1} \cline{4-4}
MLPf\_Res50\_MX & \multirow{-2}{*}{\begin{tabular}[c]{@{}c@{}}Image Classification\end{tabular}} & \multirow{-2}{*}{ResNet-50} & MXNet (NVIDIA) &  \multirow{-2}{*}{ImageNet} & \multirow{-2}{*}{Accuracy: 0.749} \\
\hline
MLPf\_SSD\_Py & & SSD (light-weight) & &   & mAP: 0.212 \\
\cline{3-3} \cline{1-1} \cline{6-6}
MLPf\_MRCNN\_Py & \multirow{-2}{*}{Object Detection} & \begin{tabular}[c]{@{}c@{}}Mask RCNN \\ (heavy-weight)\end{tabular} & \multirow{-2}{*}{PyTorch (NVIDIA)} &  \multirow{-2}{*}{\begin{tabular}[c]{@{}c@{}}Microsoft\\ COCO\end{tabular}} & \begin{tabular}[c]{@{}c@{}}Box mAP: 0.377,\\ Mask mAP: 0.339\end{tabular} \\ \hline
MLPf\_XFMR\_Py & & Transformer &  &   & BLEU score (uncased): 25 \\
\cline{3-3} \cline{1-1} \cline{6-6} 
MLPf\_GNMT\_Py & \multirow{-2}{*}{Translation} & RNN GNMT & \multirow{-2}{*}{PyTorch (NVIDIA)} & \multirow{-2}{*}{WMT17} & \begin{tabular}[c]{@{}c@{}}Sacre BLEU score\\ (uncased): 21.80\end{tabular} \\ 
\hline
MLPf\_NCF\_Py & Recommendation & \begin{tabular}[c]{@{}c@{}}Neural Collaborative\\ Filtering\end{tabular} & PyTorch (NVIDIA) & \begin{tabular}[c]{@{}c@{}}MovieLens\\ 20-million\end{tabular} & Hit rate @ 10: 0.635 \\ 
\hline
\multicolumn{6}{|c|}{\cellcolor[HTML]{ECF4FF}{\textbf{DAWNBench}}} \\ 
\hline
Dawn\_Res18\_Py & Image Classification & ResNet-18 (modified) & PyTorch (bkj) & CIFAR10 & Test accuracy: 94\%\\
\hline
Dawn\_DrQA\_Py & Question Answering & DrQA & PyTorch (Yang et al.) & SQuAD & F1 score: 0.75 \\
\hline
\multicolumn{6}{|c|}{\cellcolor[HTML]{ECF4FF}{\textbf{DeepBench}}} \\ 
\hline
Deep\_GEMM\_Cu & Dense Matrix Multiply & N/A & Bare-metal CUDA & N/A & N/A\\
\hline
Deep\_Conv\_Cu & Convolution & N/A & Bare-metal CUDA & N/A & N/A \\
\hline
Deep\_RNN\_Cu & Recurrent Layer & N/A & Bare-metal CUDA & N/A & N/A \\
\hline
Deep\_Red\_Cu & All-Reduce & N/A & Bare-metal CUDA & N/A & N/A \\
\hline
\end{tabular}%
}
\vspace{-5mm}
\label{table:summary}
\end{table*}

%% file: background.tex
In this section, we present an introduction to MLPerf~\cite{MLPerf}, DAWNBench~\cite{Coleman2017DAWNBenchA}, and DeepBench~\cite{DeepBench} benchmarks for machine learning. 

\subsection{MLPerf Benchmarks}
The MLPerf~\cite{MLPerf} benchmark suite includes workloads from image classification, object detection, translation, recommendation and reinforcement learning. 

\begin{itemize}
    \item \textbf{Image classification} is a typical deep learning application that identifies the object classes present in the image. This benchmark uses ResNet-50~\cite{Img_Recog,identity2016} model to classify images. ResNet-50 signifies a 50-layered residual network, that effectively overcomes the problem of degradation of training accuracy and is easier to optimize, and can gain accuracy from considerably increased depth. 
    
    \item \textbf{Object detection} is a technology that classifies individual objects and localizes each using a bounding box. Mask R-CNN~\cite{OBJDET} adds a branch for predicting segmentation masks on each Region of Interest (RoI), along with the existing branch for classification and bounding box regression. In Mask R-CNN, the additional mask output is distinct from the class and box outputs, as it extracts a finer spatial layout of an object. On the contrary, Single Shot Detection (SSD)~\cite{SSD} discretizes the output space of bounding boxes into a set of default boxes over different aspect ratios and scales per feature map location. The SSD model completely eliminates proposal generation and subsequent pixel or feature resampling stage and encapsulates all computation in a single network. This makes SSD easy to train and integrate into systems that require a detection component. 

    \item \textbf{Translation} is the task of converting an input text from one language to another. The model architecture - Transformer~\cite{Trans}, avoids recurrence and relies on an attention mechanism to generate global dependencies between input and output. The attention weights apply to all symbols in the sequences. On the other hand, Google's Neural Machine Translation system (GNMT) \cite{TranslationGNMT} model uses residual connections as well as attention connections. GNMT provides a decent balance between the flexibility of ``character''-delimited models and the efficiency of ``word''-delimited models, and handles translation of rare words.
    \item \textbf{Recommendation} is a task accomplished by a recommendation system, that predicts the "rating" or "preference" to an item. This benchmark uses Neural Collaborative Filtering model (NCF)~\cite{RECOMM} that can express and generalize matrix factorization under its framework. To supercharge NCF modeling with non-linearities, a multi-layer perceptron can be utilized in this model to learn the user-item interaction function.
    
    \item \textbf{Reinforcement Learning} is associated with how software agents should take actions in an environment to maximize the notion of cumulative reward. This benchmark is based on a fork of the mini-go project~\cite{Minigo}, inspired by DeepMind's AlphaGo algorithm~\cite{GOwithDNN,GOwithoutHuman}. There are four phases in this benchmark, repeated in order: selfplay, training, target evaluation, and model evaluation. Moreover, this architecture is also extended for Chess and Shogi~\cite{silver2017mastering}.
    \footnote{Since our evaluation focus is on MLPerf \texttt{v0.5} on GPU platforms, and the only GPU code of \textit{Reinforcement Learning} is the reference one, which spends more time on the CPU than the GPU, \textit{Reinforcement Learning} is excluded in the rest of the paper.}
\end{itemize}

The above mentioned MLPerf benchmarks use various datasets for training, such as ImageNet~\cite{ImageNetPaper}, Microsoft COCO ~\cite{Microsoft_COCO}, WMT17~\cite{WMT17}, and MovieLens 20-million~\cite{MovieLensPaper,MovieLens}.

Table~\ref{table:summary} displays a summary of the various workloads of MLPerf \texttt{v0.5} release, including respective models as well as the datasets used.
The metric used by MLPerf is the time taken to reach a specified accuracy or quality target, which is also listed in Table~\ref{table:summary} for each benchmark.
MLPerf benchmark implementations provided by the submitters currently include frameworks such as  PyTorch~\cite{PyTorch}, MXNet~\cite{mxnet} and TensorFlow~\cite{tensorflow2015-whitepaper}. Many of the workloads consume days of training time on powerful GPUs, as indicated in Table~\ref{table:training_time} for MLPerf's reference machine which has an NVIDIA Tesla P100 GPU.

In order to balance fairness and innovation, MLPerf takes two approaches: closed model division and open model division. The MLPerf closed model division postulates the model to be used and restricts the values of hyper parameters, such as batch size and learning rate, with the emphasis on fair comparisons of the hardware and software systems. On the contrary, in the open model division, the same problem is required to be solved using the same data set but with fewer restrictions, with the emphasis on advancing the state-of-the-art of ML \cite{MLPerf}. 

\begin{table}[htbp]
\renewcommand{\arraystretch}{1.2}
\centering
\caption{Training time of MLPerf reference implementations of the benchmark on MLPerf's reference machine (consisting one NVIDIA Tesla P100 GPU).} 
\label{table:training_time}
\begin{tabular}{|c|c|}
\hline
\rowcolor{LightBlue}
\textbf{Benchmark}                                                    & \textbf{Training Time (mins.)} \\ \hline
Image Classification & 8831.3           \\ \hline
Object Detection (SSD)  & 827.7                          \\ \hline
Object Detection (M-RCNN) & 4999.5        \\ \hline
Translation (Transformer) & 1869.8                   \\ \hline
Translation (GNMT)        & 1334.5                 \\ \hline
Recommendation (NCF)     & 46.7                          \\ \hline
\end{tabular}
\vspace{-7mm}
\end{table}

%

\input{hardware_table.tex}

\subsection{DAWNBench}
DAWNBench~\cite{Coleman2017DAWNBenchA}, developed by Stanford University in 2017, evaluates deep learning systems across different optimization strategies, model architectures, software frameworks, clouds, and hardware. It supports benchmarking of \textit{Image Classification} on CIFAR10~\cite{CIFAR10} and ImageNet~\cite{ImageNetPaper}, and \textit{Question Answering} on SQuAD~\cite{SQuAD}. DAWNBench assesses the performance based on four metrics: training time to a specified validation accuracy, cost (in USD) of training, average latency of performing inference, and the cost (in USD) of inference. It provides reference implementations and seed entries, implemented in two popular deep learning frameworks: PyTorch~\cite{PyTorch} and TensorFlow~\cite{tensorflow2015-whitepaper}. The hyperparameters that DAWNBench considers for optimizations are optimizer for gradient descent, minibatch size, and regularization.

\subsection{DeepBench}
DeepBench~\cite{DeepBench}, released in 2016, {and updated in 2017~\cite{DeepBenchUpdate},} primarily uses the neural network libraries to benchmark the performance of basic operations on different hardware. The performance characteristics of models built for various applications are different from each other. DeepBench essentially benchmarks the underlying operations such as dense matrix multiplication, convolutions, recurrent layers, and communication. For training, DeepBench specifies the minimum precision requirements as 16 and 32 bits for multiplication and addition, respectively \cite{DeepBench}. The benchmarks are written in CUDA and thus, are more fundamental than any deep learning framework or model implementation. Additionally, there is no concept of a quality target.

With research in the field of deep learning, various other benchmarks have also appeared in the past, such as Fathom~\cite{adolf2016fathom}, Training Benchmark for DNNs (TBD)\cite{TBDBenchmark}, etc., but our study is restricted to MLPerf, DAWNbench and DeepBench.



%% file: hardware_table.tex

\begin{table*}[t]
\renewcommand{\arraystretch}{1.1}
\centering
\caption{Hardware specifications of systems for experimentation.}
\label{table:hardware}
\footnotesize
\begin{tabular}{|
>{\columncolor{LightBlue}}C{1.37cm} |C{2.26cm}|C{2.26cm}|C{2.26cm}|C{2.26cm}|C{2.26cm}|C{2.26cm}|}
\hline
\textbf{Systems}           & \cellcolor{LightBlue}\textbf{T640}                                   & \cellcolor{LightBlue}\textbf{C4140 (B)}                              & \cellcolor{LightBlue}\textbf{C4140 (K)}                              & \cellcolor{LightBlue}\textbf{C4140 (M)}                              & \cellcolor{LightBlue}\textbf{R940 XA}                                & \cellcolor{LightBlue}\textbf{DSS 8440}                                  \\ \hline
\multicolumn{7}{|c|}{\cellcolor[HTML]{ECF4FF}{CPUs (\textbf{Intel Xeon Gold})}} \\ 
\hline
\textbf{Model \#}            &  6148                 & 6148                 &  6148                &  6148           &  6148          & 6142             \\ \hline
\textbf{Base freq.}            & 2.40GHz                  & 2.40GHz                  & 2.40GHz                  & 2.40GHz                  & 2.40GHz                  & 2.60GHz                  \\ \hline
\multicolumn{7}{|c|}{\cellcolor[HTML]{ECF4FF}{Memory (\textbf{Samsung/Micron DDR4})}} \\ 
\hline
\textbf{\# DIMM}          & 12                            & 12                            & 12                             & 24    & 24                      & 12                             \\ \hline
\textbf{Size}          & 16GB  & 16GB                      & 16GB                       & 16GB                          & 16GB  & 32GB        \\ \hline
\multicolumn{7}{|c|}{\cellcolor[HTML]{ECF4FF}{GPUs (\textbf{NVIDIA Tesla V100})}} \\ 
\hline
\textbf{Form Factor}            &      PCIe Full Height/Length                   &     PCIe Full Height/Length             & SXM2     & SXM2        &   PCIe Full Height/Length          &       PCIe Full Height/Length           \\ \hline
\textbf{Inter-connect} & PCIe \& UPI\footnotemark[3] & PCIe & NVLink & NVLink & UPI\footnotemark[3] & PCIe \& UPI\footnotemark[3] \\ \hline
\textbf{\# GPUs}            & 4                   & 4                 & 4             & 4                & 4                      & 8                  \\ \hline
\textbf{Memory}            & 32GB HBM2                         & 16GB HBM2                        & 16GB HBM2                   & 16GB HBM2                   & 32GB HBM2      & 16GB HBM2 \\ \hline
\multicolumn{7}{|c|}{\cellcolor[HTML]{ECF4FF}{System (\textbf{Dell PowerEdge})}} \\
\hline
\textbf{Topology}
&\includegraphics[
  height=2cm,
  trim = 148mm 0mm 62mm 0mm, 
  clip=true]{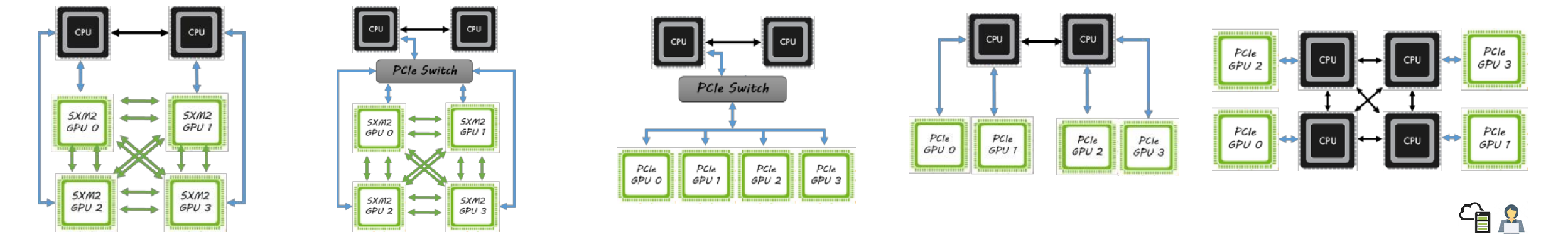}
&\includegraphics[
  height=2cm,
  trim = 98mm 0mm 112mm 0mm, 
  clip=true]{figs/topo1.ps}
&\includegraphics[
  height=2cm,
  trim = 48mm 0mm 162mm 0mm, 
  clip=true]{figs/topo1.ps}
&\includegraphics[
  height=2cm,
  trim = 3mm 0mm 207mm 0mm, 
  clip=true]{figs/topo1.ps}
&\includegraphics[
  height=1.2cm,
  trim = 196mm 10mm 3mm 0mm, 
  clip=true]{figs/topo1.ps}
&\includegraphics[
  height=2cm,
  natwidth=10in,
  natheight=5.62in,
  trim = 80mm 8mm 80mm 7mm,
  clip=true]{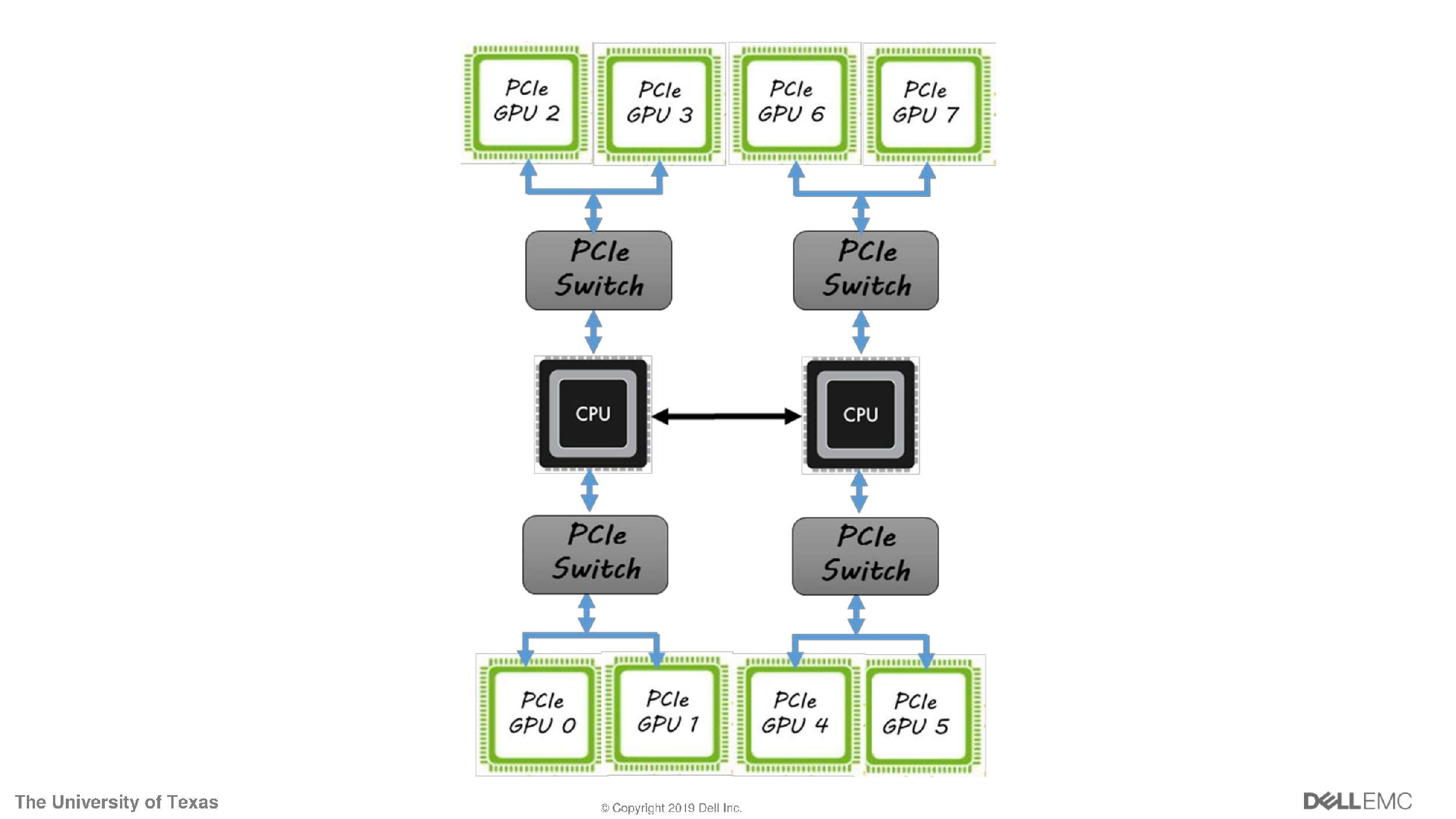}
\\ \hline
\end{tabular}%
\vspace{-3mm}
\end{table*}

%% file: methodology.tex

\subsection{System configurations}
In this work, we used different system configurations for experimentation, whose hardware specifications are highlighted in Table~\ref{table:hardware}. All the systems, except C4140 (B), operate on Ubuntu 16.04.4 LTS. The operating system on C4140 (B) is CentOS Linux 7.

\subsection{Benchmarks}\label{sec:benchmarks}
The benchmarks we chose to conduct research on are as follows:
\begin{itemize}
\setlength\itemsep{.1em}
    \item GPU submissions of the MLPerf~\cite{MLPerf} training benchmarks, which were made by Google (cloud) and Nvidia (on-premise). The submitted source codes were optimized for performance on their respective hardware. Among the various submissions, we picked Google's submission on \texttt{8x Volta V100} and NVIDIA's submission on \texttt{DGX-1} as we had access to platforms with a maximum of 8 GPUs. Note that, as there was no GPU submission for \textit{Reinforcement Learning} benchmark (one of the MLPerf training benchmarks), we exclude this benchmark from the study.
    \item From DAWNBench~\cite{Coleman2017DAWNBenchA}, for \textit{Image Classification (CIFAR10)} training we selected the ResNet-18 implementation~\cite{Py_Res18} provided by \texttt{bkj}, and for \textit{Question Answering (SQuAD)} training we chose the DrQA implementation~\cite{Py_DrQA} submitted by Yang et al.
    \item In the case of DeepBench~\cite{DeepBench}, we used four NVIDIA training benchmarks: \texttt{gemm\_bench}, \texttt{conv\_bench}, \texttt{rnn-\_bench}, and \texttt{nccl\_single\_all\_reduce}. We omitted \texttt{nccl\_mpi\_all\_reduce} as training on different nodes is not the focus of this work.
    {The aggregated numbers are used for all the kernels with different sizes, except for \texttt{rnn\_bench}, for which we only take six configurations because the benchmark takes a lot of time to profile.
    The six configurations used are Vanilla DeepSpeech (Units=1760, N=16), LSTM Machine Translation (Input=512, N=16), LSTM Language Modeling (Input=4096, N=16), LSTM Character Language Modeling (Input=256, N=16), GRU DeepSpeech (Units=2816, N=32), and GRU Speaker ID (Units=1024, N=32).}

\end{itemize}


Note that, the hyperparameters like batch size and learning rate were scaled accordingly to ensure that the run\footnote{``A  run  is  a  complete  execution  of  an  implementation  on  a  system, training a model from initialization to the quality target.'' - MLPerf~\cite{MLPerf}} completed successfully on our experimental setup.

\subsection{Measurement tools}
\par\noindent\textbf{nvprof:}
We use \texttt{nvprof} profiler from CUDA-toolkit to profile the Region of Interest (ROI) in the benchmarks.
Information collected are: invocation and duration of kernels, floating point operation counts, and memory read/write transactions.
With this information, we added data points as the representatives of machine learning workloads to the roofline plot.
\par\noindent\textbf{dstat:}
Additionally, we used \texttt{dstat}~\cite{dstat} to obtain the real-time statistics of system resource usage such as CPU usage, memory usage, disk activity, and network traffic. In UNIX platform, \texttt{dstat} gives more flexibility that combines \texttt{vmstat} (virtual memory statistics)~\cite{vmstat}, \texttt{iostat} (storage input/output statistics)~\cite{iostat}, and \texttt{netstat} (network statistics)~\cite{netstat}. The statistics can then be exported to comma-separated values for further analysis. Moreover, we can extend the functionality of \texttt{dstat} by adding plugins such as one to measure NVIDIA GPU Utilization~\cite{nvGPUutil}.
\par\noindent\textbf{dmon:}
Finally, we also make use of \texttt{dmon} which is available in Nvidia System Management Interface (\texttt{nvidia-smi})~\cite{smi} to get individual GPU usage statistics that includes GPU streaming multiprocessor usage, GPU memory usage, temperature, frequency, and PCI Express bus usage. A feature to measure the NVLink bus utilization using hardware counters is also employed in \texttt{nvidia-smi}.

\setcounter{footnote}{3}
\footnotetext{UPI: Ultra Path Interconnect}

%% file: analysis.tex
The analysis is presented on the optimized codes submitted by Google and NVIDIA to MLPerf unless specified otherwise. It may be noted from the MLPerf website that only three vendors (Google, NVIDIA, and Intel) have submitted results to MLPerf, and no vendor has submitted results for all benchmarks. The effort to run MLPerf codes on the systems mentioned in Table~\ref{table:training_time} and~\ref{table:Scalability} was non-trivial,
and some of the benchmarks are omitted from some studies due to difficulties with runs.
A statistic of kernels is available in the appendix.

\begin{figure}[!hbt]
  \begin{minipage}{0.40\textwidth}
    \centering
    \subfloat[PC1 - PC2]{
    \label{fig:PC1PC2}
    \includegraphics[
    center,
    width=0.8\textwidth,
    trim = 0mm 98mm 1mm 94mm, 
    clip=true]{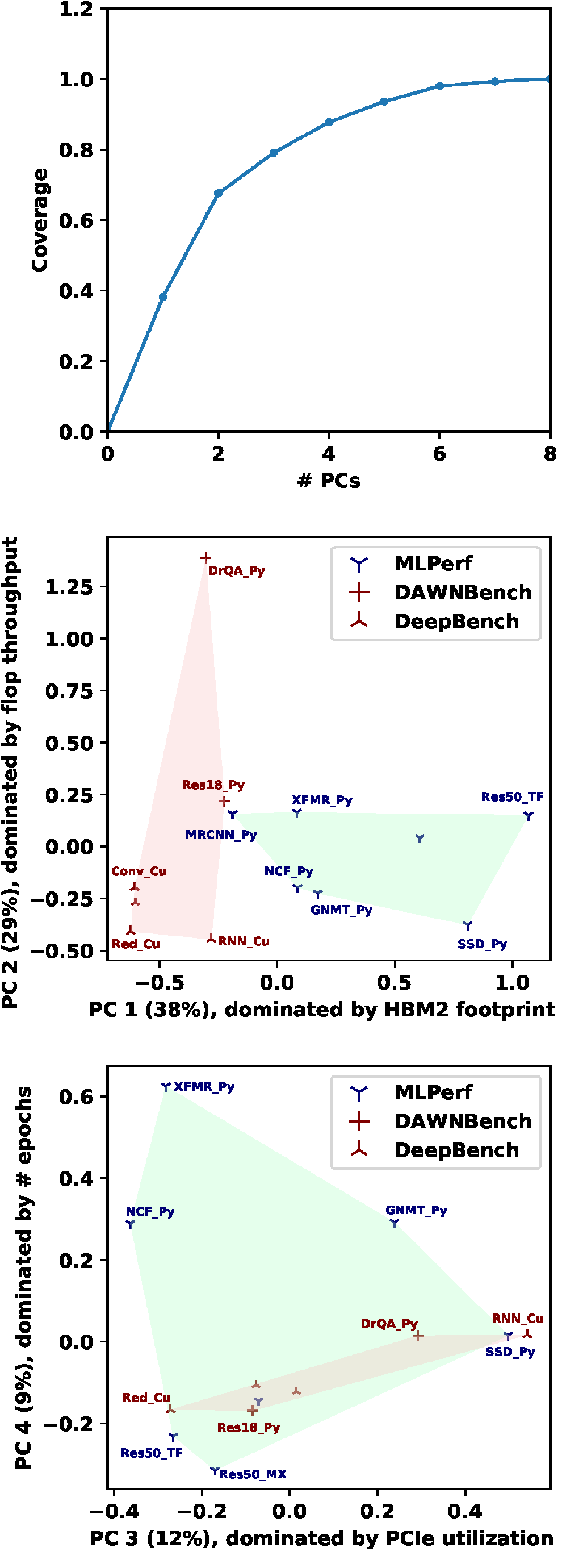}}
  \end{minipage}
  \begin{minipage}{0.40\textwidth}
    \centering
    \subfloat[PC3 - PC4]{
    \label{fig:PC3PC4}
    \includegraphics[
    center,
    width=0.8\textwidth,
    trim = 1mm 3mm 1mm 189mm, 
    clip=true]{figs/PCA_arXiv.ps}}
  \end{minipage}
  \vspace{-3mm}
  \caption{The distribution of MLPerf, DAWNBench, and DeepBench in the dominant principal component workload space.
  The dominant metric is the one with the greatest absolute value in the eigenvector of a principal component.}
  \label{fig:PCA}
  \vspace{-5mm}
\end{figure}
\begin{figure}[hbt]
  \vspace{-5mm}
  \centering
  \includegraphics[
  width=0.8\linewidth,
  natwidth=8in,
  natheight=5in,
  trim = 5mm 5mm 5mm 5mm,
  clip=true]{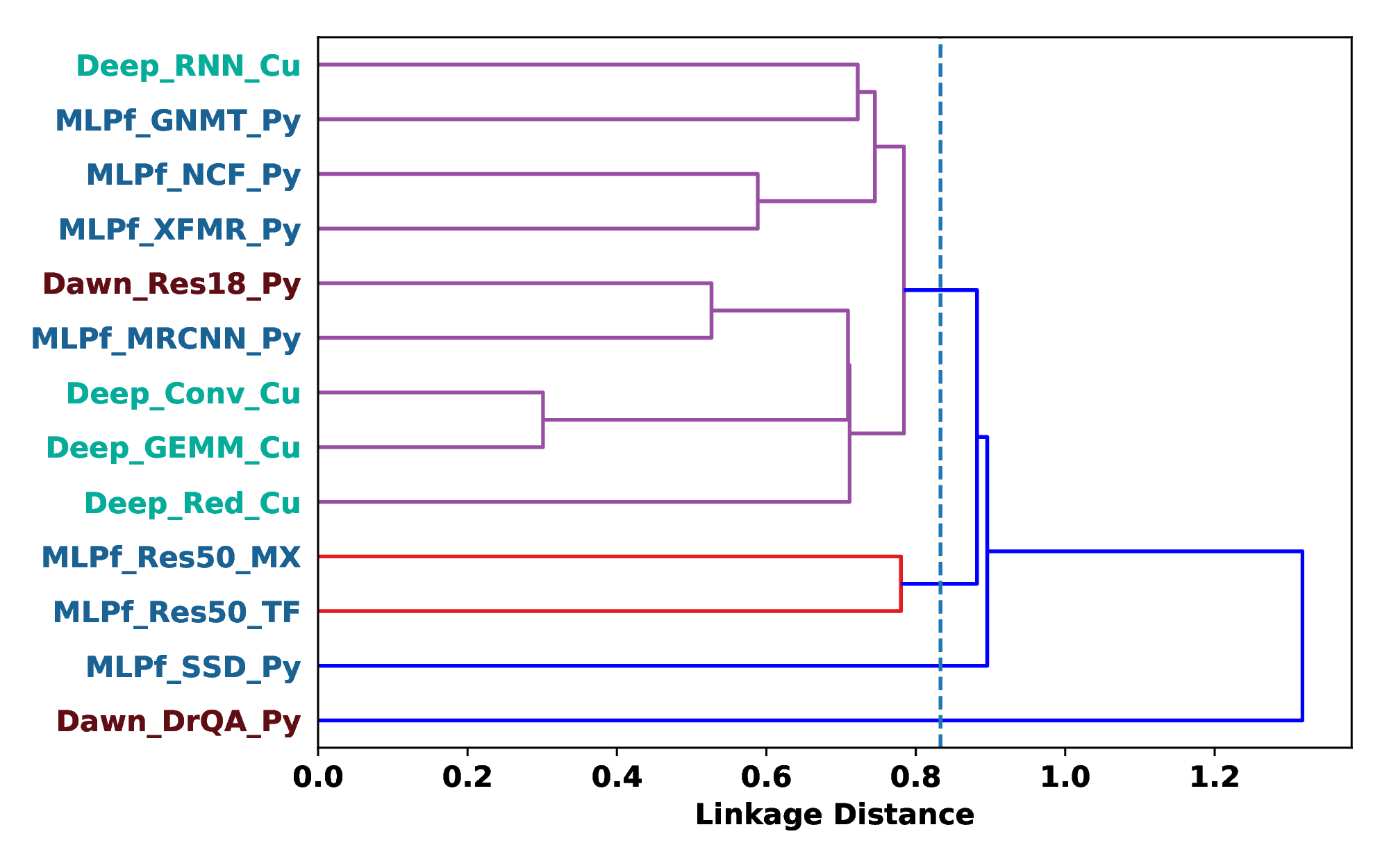}
  \caption{Dendrogram of MLPerf, DAWNBench and DeepBench benchmarks. If a subset of 4 is desired, pick one from each cluster intercepted by the vertical line at  linkage distance around 0.8.}
  \label{fig:dendrogram}
  \vspace{-5mm}
\end{figure}
\begin{figure}[hbt]
  \centering
  \includegraphics[
  width=0.95\linewidth,
  natwidth=8.5in,
  natheight=11in,
  trim = 19mm 56mm 31mm 61mm,
  clip=true]{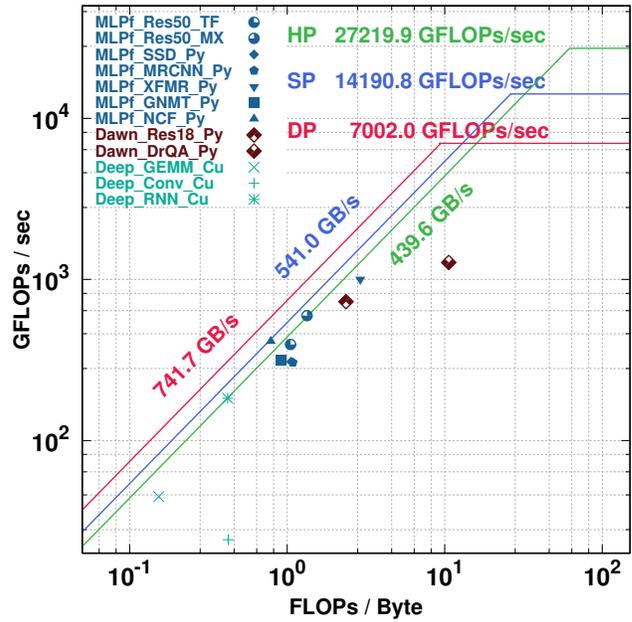}
  \caption{V100 roofline model marked with MLPerf benchmarks (labeled in blue), DAWNBench (labeled in red), and DeepBench (labeled in cyan). Red, blue, green polylines show the empirical limitations (from available memory bandwidth and computational resources) for V100 to perform double, single, and half-precision floating point operations (measured with Empirical Roofline Toolkit~\cite{ERT}).}
  \label{fig:roofline}
  \vspace{-6mm}
\end{figure}
\subsection{Similarity/Dissimilarity analysis}
\label{sec:Similarity}
\input{similarity}

\subsection{Roofline analysis}
\label{sec:roofline}
\input{roofline}

\subsection{Sensitivity of MLPerf models to Mixed Pre-cision Training {using Tensor Cores}}
\label{sec:precision}

\begin{figure}[htb]
  \vspace{-1.5mm}
  \centering
  \includegraphics[
  width=\linewidth,
  natwidth=5.6in,
  natheight=2.2in,
  trim = 0mm 1.8mm 0mm 1mm,
  clip=true]{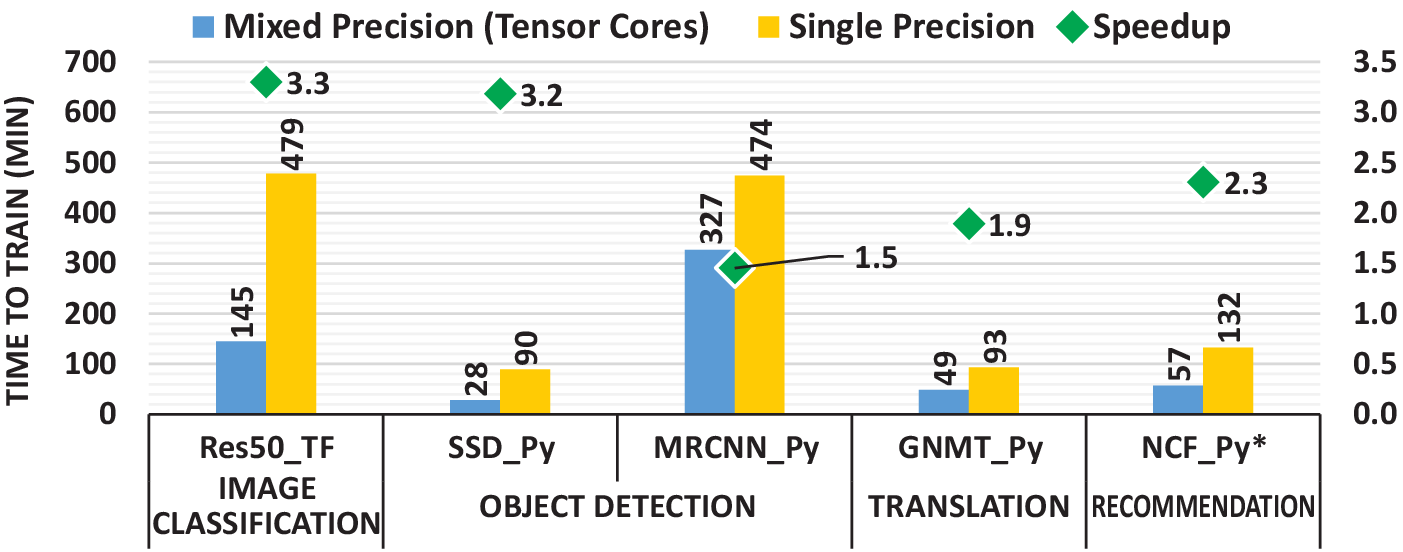}
  \caption{ Mixed Precision training {(supported by Tensor Cores) results in 1.5$\times$ to 3.3$\times$ speedups over single precision}. (Note the time of NCF\_Py is in seconds)}
  \label{fig:precision}
  \vspace{-2.5mm}
\end{figure}

Prior work (\cite{micikevicius2017mixed,hubara2016quantized,gupta2015deep} ) suggests that reduced precision training helps the deep learning training in the following ways:
\begin{itemize}
    \item Lowering the on-chip memory requirement for the neural network models.
    \item Reducing the memory bandwidth requirement by accessing less or equal bytes compared to single precision.
    \item Accelerating the math-intensive operations especially on GPUs with Tensor Cores. 
\end{itemize}
Typically, only some pieces of data employ reduced precision leading to mixed precision implementations. Moreover, employing mixed precision for training is getting easier for programmers with the release of NVIDIA's Automatic Mixed Precision (AMP)~\cite{AMP} feature on different frameworks like TensorFlow~\cite{tensorflow2015-whitepaper}, PyTorch~\cite{PyTorch} and MXNet~\cite{mxnet}. Figure~\ref{fig:precision} shows the speedup observed in different MLPerf training benchmarks by employing the use of half-precision along with single-precision when tested on DSS 8440 using 8 GPUs. The speedup observed is in the range of 1.5$\times$ in MRCNN\_Py to 3.3$\times$ in Res50\_TF.
{Thus, it can be inferred that MLPerf,  an end-to-end benchmark suite, is capable of testing the reduced precision support of processors. For example, TensorCores are tested here.}

\subsection{Compiler optimization impact on Benchmark performance}
\label{sec:compiler_optimization}

\begin{figure}[hbtp]
  \centering
  \includegraphics[
  natwidth=4.62in,
  natheight=2.25in,
  width=\linewidth
  ]{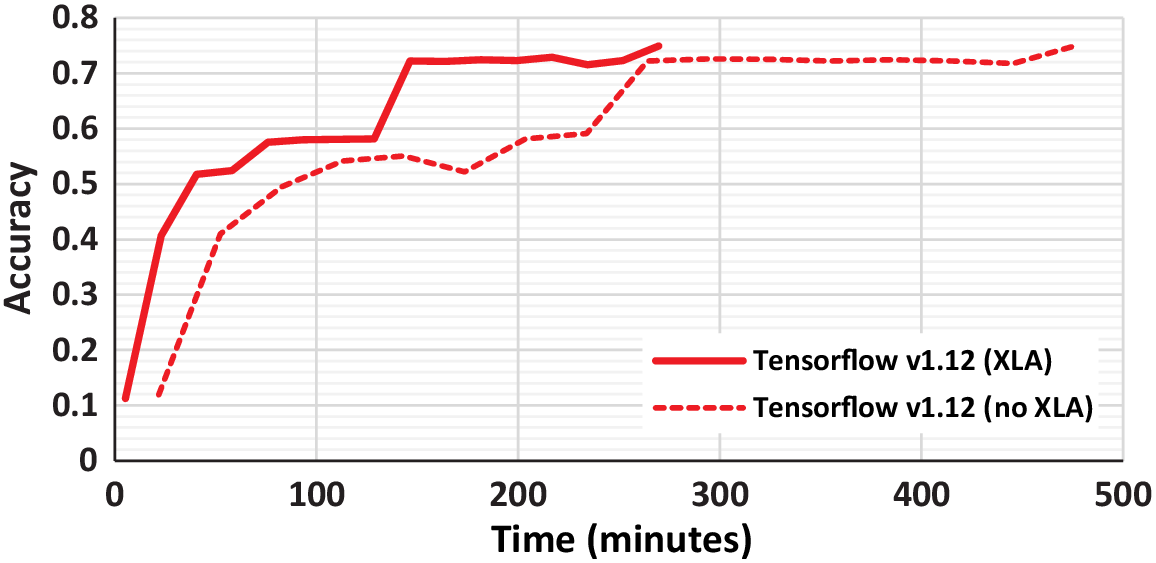}
  \caption{Image Classification {(Res50\_TF)} reaches  desired accuracy in 60\% time if compiler uses XLA optimization.}
  \label{fig:Res50_TTA_frameworks}
  \vspace{-2mm}
\end{figure}

Deep learning frameworks offer building blocks for designing, training and validating deep neural networks through a high level programming interface. They rely on GPU-accelerated libraries such as cuDNN~\cite{chetlur2014cudnn} and NCCL~\cite{nccl} to deliver high-performance for single as well as multi GPU accelerated training. In Figure~\ref{fig:Res50_TTA_frameworks}, we can see that MLPf\_Res50-\_TF takes around 270 minutes. These experiments are performed on C4140 (K) using all 4 GPUs. It is interesting to note that the TensorFlow/XLA JIT (just-in-time) compiler~\cite{XLA} optimizes TensorFlow computations and reduces the execution time by about 40\% for this use case. XLA uses JIT compilation techniques to analyze and optimize the TensorFlow subgraphs created by the user at runtime. Some optimizations are specialized for the target device. The compiler then fuses multiple operators (kernel fusion) together and generates efficient native machine code for the device. This results in the reduction of execution time and the required memory bandwidth for the application.

\begin{figure}[!hbt]
  \vspace{-5mm}
  \begin{minipage}{0.17\textwidth}
    \centering
    \subfloat[]{
    \label{fig:schedule_4GPU_all4T}
    \includegraphics[
    height=5cm,
    natwidth=7.28in,
    natheight=9.0in,
    left,
    trim = 0mm 0mm 0mm 24mm,
    clip=true]{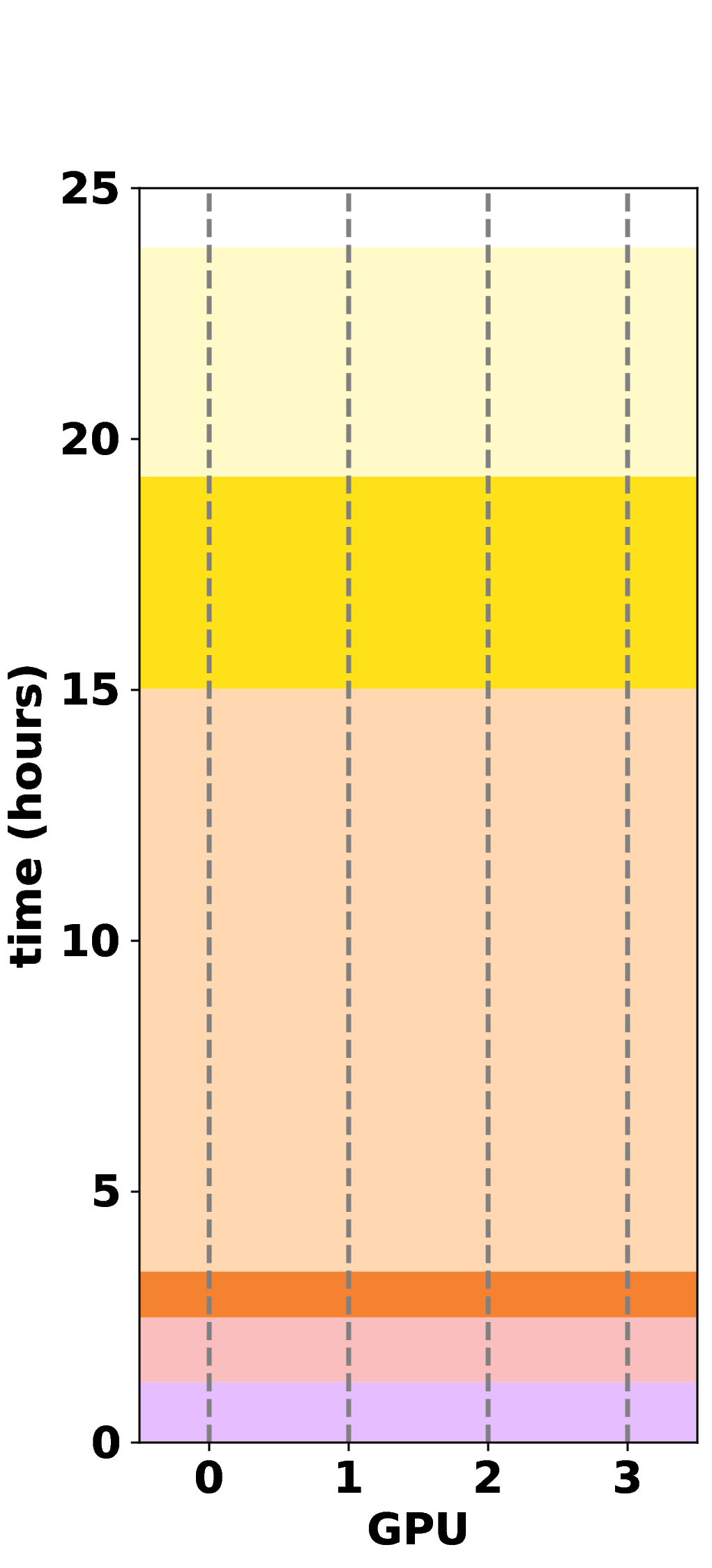}}
  \end{minipage}
  \begin{minipage}{0.17\textwidth}
    \centering
    \subfloat[]{
    \label{fig:schedule_4GPU_optimal}
    \includegraphics[
    height=5cm,
    natwidth=7.28in,
    natheight=9.0in,
    left,
    trim = 7mm 0mm 0mm 24mm,
    clip=true]{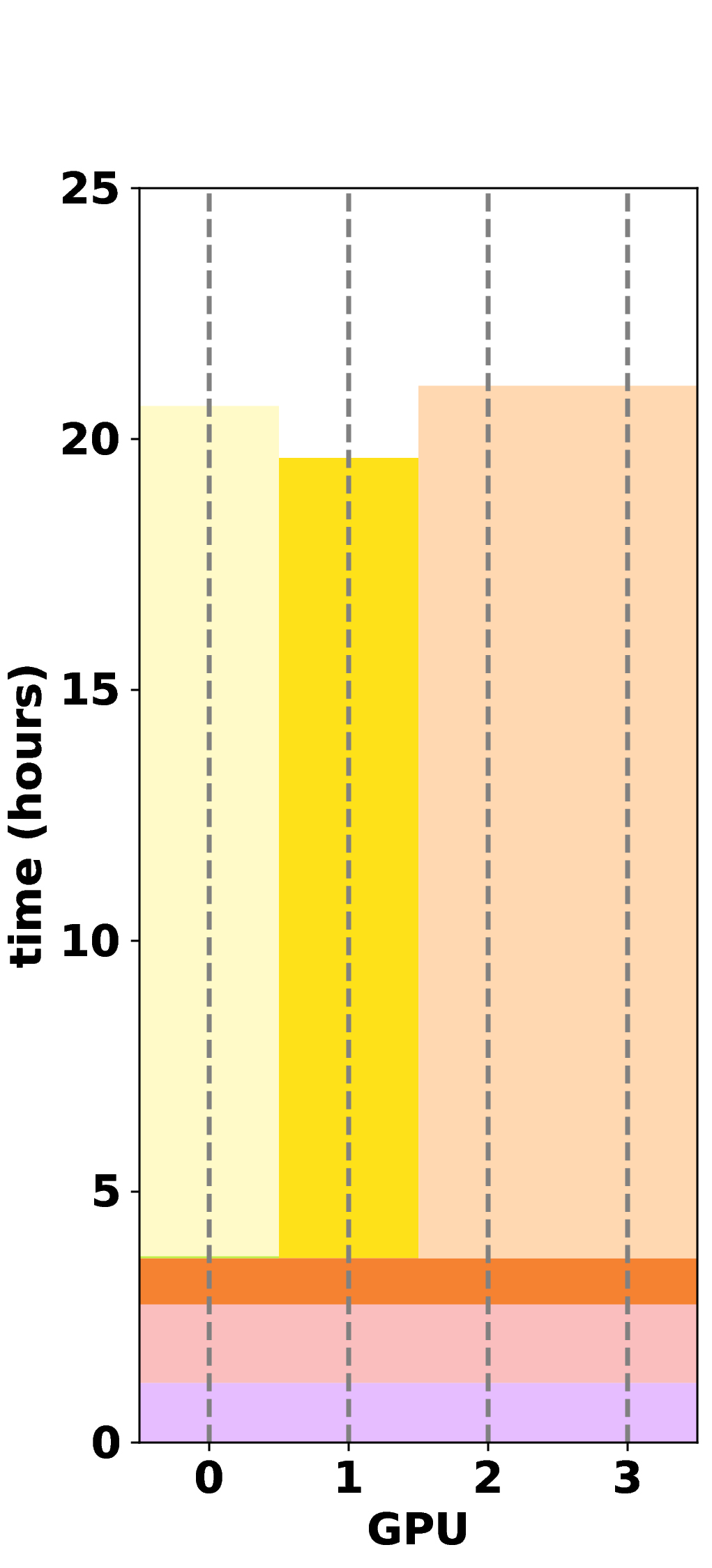}}
  \end{minipage}
  \begin{minipage}{0.1\textwidth}
    \centering
    \label{fig:schedule_legend}
    \includegraphics[
    width=0.99\textwidth,
    natwidth=1.88in,
    natheight=3.0in,
    trim = 3mm 4mm 7mm 30mm,
    clip=true]{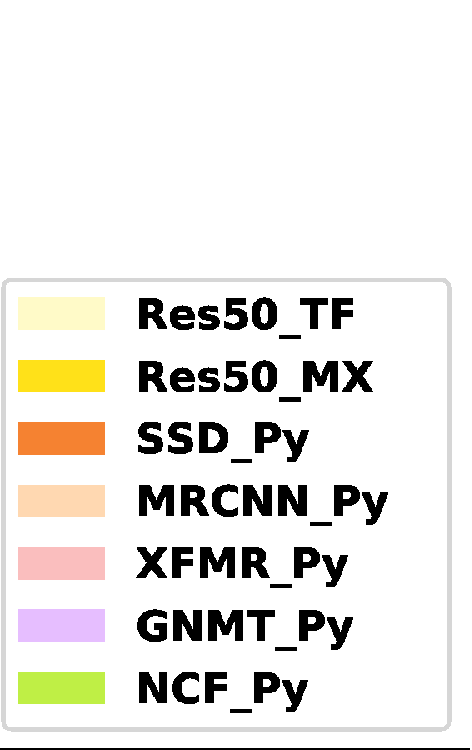}
  \end{minipage}
  \vspace{-3mm}
  \caption{Scheduling a mix of MLPerf workloads on 4 GPUs: (a) naive scheduling, which distributes one benchmark on all the GPUs one by one; (b) optimal scheduling, found by searching through the possible space, saves 2.8 hours.}
  \label{fig:schedules}
  \vspace{-1mm}
\end{figure}

\subsection{Scalability of the benchmarks}\label{sec:scalability}

\begin{table}[htbp]
\renewcommand{\arraystretch}{1.2}
\centering
\vspace{-2mm}
\caption{Scaling efficiency for distributed training.}
\label{table:Scalability}
\resizebox{0.477\textwidth}{!}{%
\begin{tabular}{|c|c|c|c|c|}
\hline
\rowcolor{LightBlue} 
\cellcolor{LightBlue} & \cellcolor{LightBlue} & \multicolumn{3}{c|}{\cellcolor{LightBlue}\textbf{Scalability (speedup)}} \\ \cline{3-5} 
\rowcolor{LightBlue} 
\multirow{-2}{*}{\cellcolor{LightBlue}\textbf{Benchmark}} & \multirow{-2}{*}{\cellcolor{LightBlue}\textbf{\begin{tabular}[c]{@{}c@{}}Training Time on\\ 1$\times$V100 (minutes)\end{tabular}}} & \textbf{1-to-2} & \textbf{1-to-4} & \textbf{1-to-8} \\ \hline
\rowcolor[HTML]{FFFFFF} 
Res50\_TF & 1016.9 & 1.92$\times$ & 3.84$\times$ & 7.04$\times$ \\ \hline
\rowcolor[HTML]{FFFFFF} 
Res50\_MX & 957.0 & 1.92$\times$ & 3.76$\times$ & 5.92$\times$ \\ \hline
\rowcolor[HTML]{FFFFFF} 
SSD\_Py & 206.1 & 1.94$\times$ & 3.72$\times$ & 7.28$\times$ \\ \hline
\rowcolor[HTML]{FFFFFF} 
MRCNN\_Py & 1840.4 & 1.76$\times$ & 2.64$\times$ & 5.60$\times$ \\ \hline
\rowcolor[HTML]{FFFFFF} 
XFMR\_Py & 636.0 & 1.42$\times$ & 2.92$\times$ & 5.60$\times$ \\ \hline
\rowcolor[HTML]{FFFFFF} 
NCF\_Py & 2.2 & 1.88$\times$ & 2.16$\times$ & 2.32$\times$ \\ \hline
\end{tabular}%
}
\vspace{-5mm}
\end{table}

The scalability study is performed on a system with 8 GPUs, the DSS 8440,
where the number of GPUs employed to train the model is controlled.
Ideally performance speedup of using 2 GPUs, 4 GPUs and 8 GPUs over 1 GPU should be 2$\times$, 4$\times$ and 8$\times$, respectively.
Table~\ref{table:Scalability} shows the scalability trends for every MLPerf benchmark except for GNMT\_Py.
The training time using a single GPU is also added to provide a better understanding.
We can see that some of the benchmarks like Res50\_TF, Res50\_MX, and SSD\_Py scale well with the number of GPUs while for others increasing the number of GPUs beyond a certain point is not rewarding enough.
For instance, in the case of Res50\_TF, when the number of GPUs is increased from 1-to-2, from 1-to-4, and from 1-to-8, training time improves by approximately 1.9$\times$, 3.8$\times$, and 7$\times$, respectively.
On the contrary, for NCF\_Py the speedup achieved over a single GPU is 1.9$\times$, 2.2$\times$, and 2.3$\times$ when the number of GPUs is increased to 2, 4, and 8, respectively.
This data does not justify increasing the number of GPUs beyond 2 for training \textit{Recommendation} benchmark.
We believe the small dataset (MovieLens 20-million) causes this behavior for the benchmark.
Small dataset limits the maximum batch size which as a result restricts the scalability of the benchmark.
Few other benchmarks, such as XFMR\_Py and MRCNN\_Py fall between the most and the least scalable ones, providing a scale-up by a factor of roughly 1.6$\times$, 2.8$\times$, and 5.6$\times$ for 2, 4, and 8 GPUs, respectively.


Such differences in scalability between different workloads give users hints to schedule a mix of machine learning training tasks.
The naive scheduling scheme,
that sequentially distribute every workload to all resources at once,
avoids fragmentation, and keeps the resources busy all the time.
However, it may not be the most efficient way in terms of total training time,
because users having multiple GPUs can choose to distribute some scalable workloads,
while they decide to run workloads with poor scalability in sets simultaneously on fewer GPUs.
Thus, the system administrators associated with super computing clusters might be interested in finding an effective algorithm to schedule various of machine learning training jobs submitted from researchers, developers, and all other kinds of machine learning users.
To show the potential benefit, we search through all permutations of scheduling 7 MLPerf benchmarks on multiple GPUs,
and Figure~\ref{fig:schedules} presents 4-GPU scheduling for illustration.
In each subfigure,
available GPUs are listed along the x-axis,
with vertical dashed lines as their timelines.
Different color shades under the timeline correspond to the executions of the 7 different MLPerf workloads.
{Figure~\ref{fig:schedule_4GPU_optimal} shows the shortest scheduling of the 7 MLPerf benchmarks on 4 GPUs.
Compared with the naive scheduling in Figure~\ref{fig:schedule_4GPU_all4T},
it saves about 2.8 hours to finish all the training tasks.}
In the optimal scheduling, the workloads chosen to be distributed on 4 GPUs,
namely XFMR\_Py and SSD\_Py are the scalable benchmarks we observed above.
MRCNN\_Py gets two GPUs to execute due to its medium scalability.
Two \textit{Image Classification} workloads, Res50\_MX and Res50\_TF, are assigned to single GPUs separately to achieve faster training time. Note that, two similar workloads running in parallel will provide lower training time than running them in a distributed fashion even if they are highly scalable.
{Similarly, optimal scheduling could save around 4.1 hours and 0.4 hours for 2-GPU and 8-GPU settings, respectively.
It is worth mentioning that this performance gain is without any effort in optimizing the software or adding  costly hardware.}

%% file: similarity.tex
{We perform Principal Component Analysis (PCA) on 8 collected workload characteristics
(namely, PCIe utilization, GPU utilization, CPU utilization, DDR memory footprint, HBM2 footprint, flop throughput, memory throughput, and number of epochs),}
and visualize the distribution of the targeted machine learning benchmarks in the workload space. This analysis helps us to understand how similar and different these benchmarks are.
In addition, we generate the dendrogram in order to help users to pick the most representative benchmarks of certain number according to their time budget.



As shown in Figure~\ref{fig:PC1PC2},
MLPerf benchmarks are so different from DeepBench kernels as well as DAWNBench benchmarks on PC1,
that they become two isolated clusters (with outliers labeled) sitting in two sides.
PC1 is dominated by {GPU memory} footprint. The location in the space is actually a reflection of the fact that DeepBench kernels and DAWNBench benchmarks are working on relatively smaller datasets,
and they cannot stress {GPU memory} as much as MLPerf benchmarks can.
On the PC2 axis MLPerf benchmarks have a shorter span than other benchmark do,
mainly because MLPerf benchmarks are optimized end-to-end applications,
having a stable floating point operation throughput,
while while more diversity exists in the other benchmarks
(e.g., the communication kernel Deep\_Red\_Cu even has zero floating point operations).
MLPerf benchmarks are more sparsely-spread on the PC3-PC4 plane (Figure~\ref{fig:PC3PC4}),
and cover what other benchmarks cover.
The intra-suite diversity is exposed in Figure~\ref{fig:PCA} as well.
For PC1 to PC4 (covering 88\% variance), each MLPerf benchmark gets at least one chance to extend the boundary,
and there are no two MLPerf benchmarks that are very close to each other.

The dendrogram shown in Figure~\ref{fig:dendrogram} presents the result of linkage-distance-based hierarchical clustering,
where each benchmark starts as a leaf node then the two benchmarks closest to each other (i.e., most similar) are linked first,
for instance, MLPf\_Res50\_TF and MLPf\_Res50\_MX.
Dendrogram is more useful than presenting the similarities between benchmarks,
it facilitates the benchmarks selections for users who do not want or cannot run all the benchmarks due to time or cost limitation.
For example,
in Figure~\ref{fig:dendrogram},
the dashed line crossing 4 vertical lines filters out 4 most representative subsets for people can only do evaluation with 4 benchmarks.
The user is supposed to use Dawn\_DrQA\_Py, MLPf\_SSD\_Py,
one of MLPf\_Res50\_TF and MLPf\_Res50-\_MX {(take MLPf\_Res50\_TF for example)},
and another from the purple group (with all the benchmarks left, {take Deep\_Red-\_Cu for example}).
{As a validation for the subsetting, we report the range of 8 metrics covered by the 4 selected benchmarks with respect to all:
PCIe utilization 0.3\%\textasciitilde100\%, GPU utilization 0\%\textasciitilde95.6\%, CPU utilization 0\%\textasciitilde100\%, DDR memory footprint 0\%\textasciitilde100\%, HBM2 footprint 0\%\textasciitilde100\%, flop throughput 0\%\textasciitilde100\%, memory throughput 8.0\%\textasciitilde100\%, and number of epochs 0\%\textasciitilde98.4\%.}


%% file: roofline.tex
A roofline model~\cite{roofline} is a visual representation of the maximum attainable performance for a given workload in a given hardware by combining the processing core performance, memory bandwidth, and the data locality.

Figure~\ref{fig:roofline} presents the roofline model for a single Tesla V100 GPU and machine learning workloads we studied.
The runs were carried out on the T640 system, invoking just one GPU.
The vertical axis represents the compute capability that can be expressed usually in a unit of Floating Point Operation per Second (FLOPS/sec).
Meanwhile, the horizontal axis denotes the arithmetic intensity,
which is the ratio between floating point operations that can be performed per unit data using Floating Point Operations per Byte (FLOPs/Byte) as the unit.
Memory-bound workloads have lower arithmetic intensity,
hence their performance are limited by memory bandwidth (corresponding to the slope of the slash lines in Figure~\ref{fig:roofline}).
Compute-bound workloads have high enough arithmetic intensities so their performance are limited by the computational resources (corresponding to the horizontal lines in Figure~\ref{fig:roofline}).
We indicate the location of different machine learning workloads with points in different shapes.
Workloads from the same benchmark suite are assigned the same color.
As we can see from the Figure~\ref{fig:roofline},
MLPerf benchmarks are more optimized than DeepBench kernels so that there is more data reuse, achieving higher arithmetic intensity,
while the two DAWNBench workloads shows even higher arithmetic intensities with higher throughputs.
Nevertheless, all the workloads are memory-bound
(have not cross the turn point, and touch the horizontal lines).
This observation implies memory is the system bottleneck for machine learning workloads, and we should dedicate more resources to memory interface for a well-balanced system.

%% file: sys_measurements.tex
In this section we present observations made on system level utilization measurements that were performed using the measurement tools such as \texttt{dstat} and \texttt{dmon} to better understand the impact of running DL training workloads and system requirements for the different models. The experimentation is performed on C4140 (K) system by appropriately regulating the number of GPUs. 
\vspace{+10pt}

\input{util_table} 

\subsection{CPU utilization across different workloads}

In the previous section, we presented the scalability of each benchmark for one, two, four, and eight GPUs runs. Although most of the computation of the MLPerf benchmark submissions in this paper are offloaded to the GPUs, it will be interesting to see how the CPU is utilized during the execution of the benchmarks as we increase the number of GPUs. We run each workload on C4140 (K) platform and configure accordingly to use one, two, or all four GPUs available on that platform. We monitor the CPU usage periodically using \texttt{dstat}. 

The average CPU usage while running one, two, and four GPUs is summarized in Table~\ref{table:utilization}. Note that, the average CPU usage includes the operating system (e.g., kernel, low-level driver) usage as well as that used by the user programs. In general, as we double the number of GPUs that are used to run each workload, the utilization of the CPU roughly doubles. This trend is observable for all submissions to MLPerf which indicates that the CPU must have adequate performance to keep all GPUs busy otherwise it will become a bottleneck during the run.

Among the MLPerf submissions, MLPf\_Res50\_TF has the highest CPU Utilization followed by MLPf\_Res50\_MX. This is because, compared to other workloads, both \textit{Image Classification} benchmarks require CPU to perform more packaging of the data before dispatching them to the GPUs and post-processing the data after the GPUs finish the requested tasks. Moreover, the dataset used for \textit{Image Classification} benchmark is significantly bigger (around 300GB) compared to datasets for other benchmarks. Since it is not feasible to store such a big chunk of data on GPU memory, the CPU has to coordinate small parts of the dataset that can be stored in GPU memory at one time. The GPU can then perform a partial computation. This copying back and forth between CPU memory and GPU memory also increases the utilization of CPU. MLPf\_NCF\_Py shows lowest CPU utilization followed by MLPf\_GNMT\_Py and MLPf\_XFMR\_Py. The Object Detection workloads are in the middle in terms of CPU utilization. 

Another observation that we would like to highlight comes from Dawn\_DrQA\_Py. Although this benchmark runs on a single GPU, it has the highest CPU usage of all the workloads included in the Table~\ref{table:utilization}. Unfortunately, this benchmark also shows least GPU utilization among all the workloads, around 20\%, which indicates that a major part of the computation is performed on the CPU with few tasks that can be offloaded to the GPU.  

\subsection{GPU utilization for different workloads}

The GPU utilization as given in Table \ref{table:utilization} is the sum of the utilization of every GPU that is used during the runtime. Therefore, single-, dual-, and quad-GPU run will have maximum utilization of 100\%, 200\%, and 400\%, respectively. For \textit{Image Classification} workloads, both MLPf\_Res50\_TF and MLPf\_Res50\_MX, show near identical GPU utilization with around 85\% GPU usage for single-GPU run, around 190\% GPU usage (i.e., around 95\% utilization per GPU) for dual-GPU run, and around 375\% GPU usage (i.e., around 93.5\% utilization per GPU) for quad-GPU run. 

Most of the submissions to MLPerf show a similar trend for single-GPU and dual-GPU runs. Moreover, MLPf\_NCF\_Py shows decreasing individual GPU usage for quad-GPU run compared to dual-GPU run. This observation agrees with the one mentioned in Section~\ref{sec:scalability} that due to the limited scope of increase in the batch size for the workload, it is unable to utilize the GPUs efficiently. The increasing of communication cost for multi-GPU run that can impact individual GPU utilization is confirmed by Deep\_Red\_Cu benchmark from DeepBench which shows the same trend. 

\subsection{CPU and GPU memory footprint}
The system memory is mostly used to store the dataset that is used for the training as well as the intermediate data required between computations. In the case when the dataset is too large to fit in the GPU memory, the system memory acts as a buffer to store the dataset. The user program will move the data back and forth between the system and GPU memory to perform partial calculations. Moreover, in an extreme case, the dataset can be too large to be stored inside the system memory. Thus the disk storage (e.g., hard disk drive, solid state drive) is used to store them, and the CPU is responsible for coordinating the switching between each part of the dataset.

From Table \ref{table:utilization}, we can notice that the system memory footprint roughly doubles every time we double the number of GPUs. The GPU memory footprint is the total memory footprint for every GPU used during the run. Note that, the footprint of GPU memory depends on the batch size, and the batch sizes for the experiments are scaled accordingly from the original submissions as mentioned in Section~\ref{sec:benchmarks}. 


Although the table only shows the memory footprint of each benchmark, we would like to emphasize that the heterogeneity of the medium where the dataset is stored may become a bottleneck especially for memory-bounded applications which perform data exchange frequently. In this case, the interconnect bandwidth between each storage medium and the intelligence of the program to overlap the data transfer just before the next computation and to manage the locality of the data can play a crucial factor. 

In our C4140 (K) platform, for example, each CPU has 96GB of memory consisting six 16GB DDR4-2666 DIMMs in hexa-channel memory configuration. The theoretical unidirectional memory bandwidth available to each CPU is around 128GBps~\cite{CPUBandwidth}
while the Intel's proprietary Ultra Path Interconnect (UPI) that links two CPUs has only unidirectional theoretical bandwidth of 20.8GBps~\cite{UPIBandwidth}.
In a case when a CPU needs a part of the dataset stored in other CPU's memory, the performance of data transfer will be significantly reduced (i.e., 128GBps direct access for local DRAM v.s. 20.8GBps neighbor DRAM access via UPI). 

The same thing happens with a GPU that has more limited dedicated memory. In our C4140 (K) platform, each Nvidia Tesla V100 is equipped with 16GB HBM2 stacked memory which is capable of 450GB/s unidirectional bandwidth. In the case that the dataset cannot be fully stored inside the GPU memory, the CPU should bring a part of the dataset from the system memory into the GPU memory. This data exchange uses PCIe 3.0 bus which connects the CPU and GPU and able to provide theoretical unidirectional bandwidth of 15.8GBps for x16 lanes which limits the performance of data transfer.


\begin{figure*}[hbt]
  \centering
  \includegraphics[
  width=\linewidth,
  natwidth=9.14in,
  natheight=2.38in,
  clip=true]{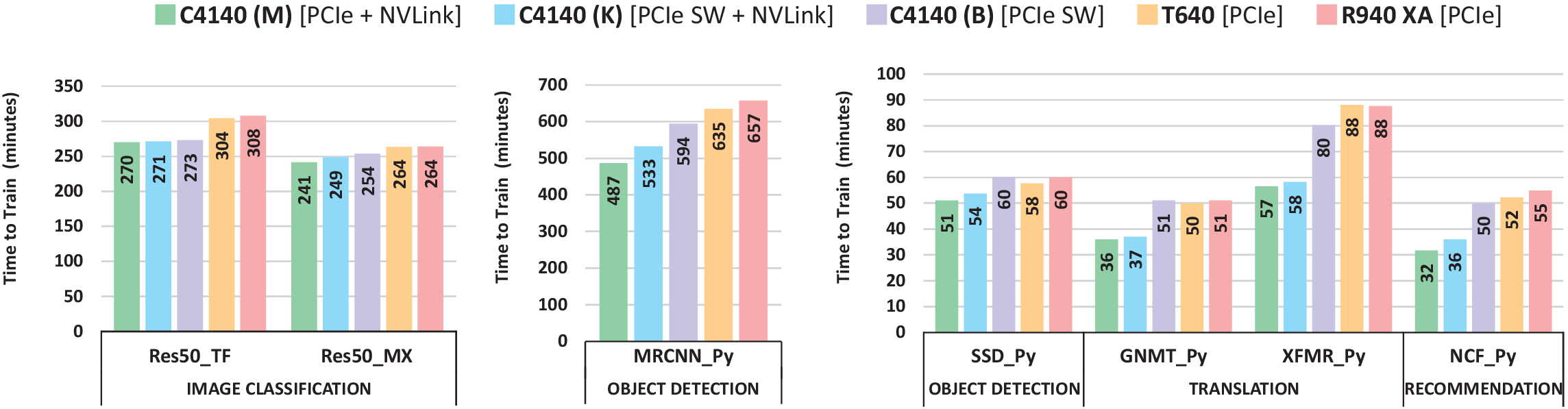}
  \caption{Training time on 4-GPU systems. Time on systems with NVLink interconnect (the first 2 bars) is less than training time on the remaining systems. (Note that, the time of NCF\_Py is in seconds)
  }
  \label{fig:Topology_comp}
  \vspace{-4mm}
\end{figure*}

\subsection{System and GPU bus utilization}
In the previous section, we have mentioned that interconnection bus between CPU-GPU and between GPU-GPU may play an important role in determining the overall system performance. Moreover, as we have learnt previously, the choice  interconnection topology between CPU and GPU should be considered carefully. In this section, we will explain more details about how the performance is impacted by the interconnection bus based on the data on Table \ref{table:utilization}.

Modern microprocessor systems use PCI Express (PCIe) bus as the interconnection standard between CPU and external peripheral that requires high-speed data communication. PCIe 3.0 standard, introduced in 2010, has been widely adopted by most computer system products available in today's market. PCIe 3.0 provides theoretical unidirectional bandwidth up-to 984.6 MBps per lane and up-to 15.8 GBps per PCIe 3.0 compatible device connected using 16 PCIe 3.0 lanes (PCIe 3.0 x16). This massive bandwidth, in theory, should be sufficient for most of the peripheral devices including GPU, network interface card, and non-volatile memory storage. 

Usually a GPU is connected to the CPU using PCIe 3.0 x16 to assure that there is plenty of bandwidth between them. High bandwidth is easy to achieve for a single-GPU system, but more complicated for a multi-GPU system since the number of PCIe 3.0 lanes that the CPU has are limited. High-end Intel Xeon may have up to 48 lanes of PCIe 3.0 which are then allocated to various devices. With this constraint, each GPU on a four GPUs system, for example, may only be assigned eight PCIe 3.0 lanes. While it depends on how we use the GPU and how intense the data exchange happens between the CPU and GPU, some applications like gaming may find PCIe 3.0 x8 already provides plenty of bandwidth. On the other hand, this much bandwidth may not be optimal for deep learning training.  

Alternatively, a PCIe switch, such as those manufactured by PLX Technology, can be used to provide additional PCIe lanes; thus each GPU can have PCIe 3.0 x16 lanes. This switch will be useful for GPU-to-GPU communication since the data exchange will only take place on the switch without going over to the CPU. However, the switch will not be beneficial to improve the bandwidth between CPU and all GPUs on the system as the effective CPU-to-GPU bandwidth is still limited by what the CPU has. We will discuss the interconnection topology in detail and how it affect the performance in Section~\ref{sec:topologyimpact}.

Furthermore, apart from CPU-to-GPU communication, PCIe bus can be used for GPU-to-GPU communication for a multi-GPU system. Although each GPU can be allocated with PCIe 3.0 x16 lanes, the available bandwidth may not be sufficient for some workloads that require intensive data exchange between the GPUs. Therefore, an additional bus has been developed to be used specifically for GPU-to-GPU communication such as NVLink which is high-speed proprietary interconnect system in NVIDIA GPUs. Each NVLink lane provides 25 GBps theoretical unidirectional bandwidth. The Nvidia Tesla V100 GPU in SXM2 form factor has six NVLink lanes which are capable of transferring data with theoretical unidirectional bandwidth of 150GBps. This is significantly faster than what PCIe 3.0 x16 can offer.


Table~\ref{table:utilization} shows the PCIe 3.0 bus utilization between CPU and GPU available on the system as well as NVLink utilization between GPU and GPU. The value presented in the table is the sum of PCIe 3.0 bidirectional PCIe bus utilization for each GPU that is used during the run, and the sum of NVLink lane utilization from each GPU used during the run. As we can see from the table, the data transfer rate over NVLink bus increases as we add more GPU for the run. The Deep\_Red\_Cu and the MLPf\_NCF\_Py use the highest bandwidth of NVLink which means that the data exchanges between GPU for those benchmark are intensive. On the other hand, the utilization of PCIe 3.0 bus increases as we add more GPU which is as expected. In a multi-GPU system equipped with NVLink, the PCIe 3.0 bus is used only for communication between CPU and each GPU because the GPU to GPU communication has been offloaded into the higher speed NVLink.

\begin{figure*}[hbt]
  \centering
 \includegraphics[
  width=\linewidth,
  natwidth=10.2in,
  natheight=3.22in,
  clip=true]{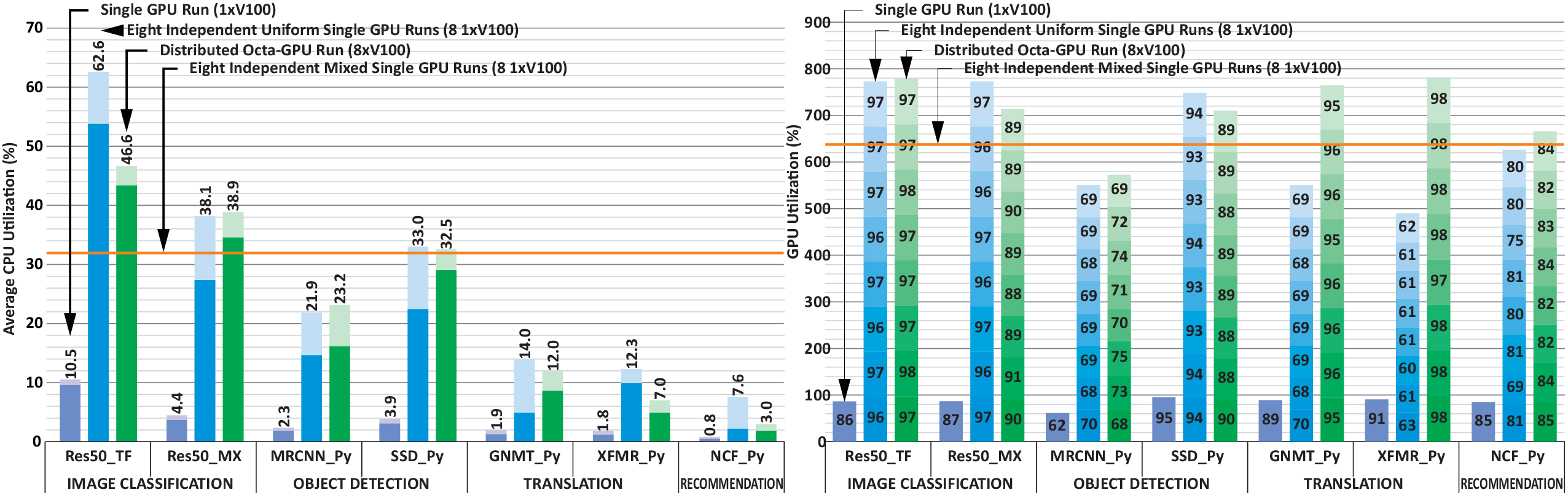}
  \caption{Utilization of CPU (left) and GPU (right) for single-GPU run, 8 independent uniform single-GPU runs, distributed 8-GPU run, and 8 independent mixed single-GPU runs. Higher CPU utilization in multiple-run vs distributed in 5 of 7 cases.}
  \label{fig:Mixed_Run_CPU_GPU_Utilization}
 \vspace{-4mm}
\end{figure*}
\subsection{Impact of GPU-Interconnect Topology}
\label{sec:topologyimpact}

To reduce the training time it is becoming increasingly common to scale deep learning (DL) training across multiple GPUs within a system. There are multiple ways in how GPUs can be connected within the system. Primarily there are two options available - using a PCIe based interconnect (which may include PCIe switches if the number of lanes from the CPU is not sufficient) and using NVIDIA's proprietary interconnect like NVLink. The theoretical bandwidth of an NVLink interconnect is 10$\times$ higher than PCIe (300 GB/s vs. 32GB/s)~\cite{NVIDIA_TESLA_V100}.
Additionally, communication libraries like NCCL from NVIDIA are optimized to perform GPUDirect peer-to-peer (P2P) direct access when NVLink is available between GPUs, which can lower training times if there is significant peer-to-peer communication during model training. GPUDirect P2P is also feasible in certain PCIe topology designs where GPUs are the same PCIe domain (single root complex). Using MLPerf, we conduct a performance evaluation of five different 4-GPU platforms, each of them with a unique GPU interconnect topology. Table~\ref{table:hardware} shows how the GPUs are interconnected for the servers used in this study.  

Two of the five servers, C4140 (M) and C4104 (K) include the high-speed proprietary NVLink interconnect to provide 100GB/s bandwidth between any two GPUs. The difference between the two NVLink based designs is the use of a PCIe switch in the C4140 (K) to aggregate the PCIe connections to the GPUs. The remaining three systems use PCIe based interconnects. They use very different approaches in how the GPUs are connected to the CPUs and in how they communicate with other GPUs. One system C4140 (B), uses a 96-lane PCIe switch that allows for 4 GPUs to be hosted in a single PCIe domain where it can perform GPUDirect peer-to-peer (P2P) between the GPUs using the PCIe switch. This is not feasible in the other two PCIe based interconnect platforms - the T640, where two GPUs are hosted per CPU and R940 XA which is a 4 CPU platform with each GPU connected directly using the PCIe lanes of the CPU.

The training times for the different servers are plotted in Figure~\ref{fig:Topology_comp} which illustrates the impact of GPU interconnect topology on DL training times. As expected, due to lack of GPUDirect P2P capability between any of the GPUs, the T640 and R940 XA take the longest time to train all the MLPerf models. Conversely, the two servers that use NVLink interconnect (the C4140 (M) and (K) systems) show the best training times across all the MLPerf models. However, the performance improvements differ depending on the model that is being trained and ranges from 42\% and 17\% for the \textit{Translation} benchmarks, 30\% for MLPf\_MRCNN\_Py to 11\% for the \textit{Image Classification} benchmarks. The C4140 (M) which uses a PCIe topology, but can perform GPUDirect P2P between GPUs due to all GPUs connected to a PCIe switch, shows performance parity to the NVLink platform for the \textit{Image Classification} benchmarks and better performance than the R940 XA and T640 servers for remaining benchmarks. This platform provides a mix of flexibility that is available when using PCIe based GPU cards in addition to higher performance over PCIe based designs that do not support GPUDirect P2P transactions between GPUs.

\subsection{Impact of job types on system utilization}
{GPU infrastructure in cloud or on-premise data centers typically hosts different classes of training jobs with different purposes:

\textbf{Distributed-run}: to train a large complex model over a large training dataset across multiple GPUs for fast time-to-solution.

\textbf{Multiple-run}: to sweep hyper-parameter space of the same model, typically having one training run with different settings of hyper-parameters on each GPU on the same test dataset.
 
 \textbf{Mixed-run}: different users submit different jobs that are training smaller models using single GPU each on a cluster

In this section, we compare the system resource utilization of these three methods of running machine learning workloads on a multi-GPU system (the 8-GPU DSS 8440).
}


Figure~\ref{fig:Mixed_Run_CPU_GPU_Utilization} shows the CPU and GPU utilization for each method. In general, running multiple instances of the same benchmark requires higher CPU utilization compared to a single instance on multiple GPUs (distributed-run). It is because for multiple-run, each instance has its own host (CPU) program that performs pre-processing, controls the GPU computation, and collects the computation from the GPU. Thus, CPU is required to handle each host program, hence, leading to higher CPU utilization. On the other hand, the mixed-run CPU utilization is roughly the same as the sum of CPU utilizations of a single GPU run for each workload.  

In GPU utilization,  the topology of how the GPUs are connected to the CPU plays an important role. MLPf\_NCF\_Py, MLPf\_XFMR\_Py, GNMT\_Py, MLPf\_MRCNN\_Py, and the MLPf\_Res50\_TF have higher GPU utilization for the distributed eight-GPU run compared to eight independent uniform single GPU runs. It turns out that their usage of PCIe bus for the distributed run is significantly higher compared to the independent run. During the distributed-run, total data transfer rate for MLPf\_NCF\_Py, MLPf\_XFMR\_Py, MLPf\_-GNMT\_Py, MLPf\_MRCNN\_Py, and MLPf\_Res50\_TF over PCIe bus reaches 58.95 GBps, 51.90 GBps, 39.1 GBps, 19.51 GBps, and 16.97 GBps, respectively. Meanwhile, for the multiple-run, they only use 276 MBps, 606 MBps, 2.07 GBps, 2.07 GBps, and 12.39 GBps, respectively. We suspect that most of GPU utilization is coming from communicating between GPUs and as there is no NVLink for GPU-GPU communication on this system. Each GPU will compete for the PCIe bus as well as for the UPI link. On the other hand, the MLPf\_Res50\_MX and MLPf\_SSD\_Py have the opposite behavior. Here the total data transfer rate for the distributed-run is smaller than the multiple-run.

%% file: util_table.tex
\begin{table}[!hb]
\renewcommand{\arraystretch}{1.1}
\vspace{-5mm}
\centering
\caption{System resource usage statistics on  C4140 (K). Utilization and footprint increase with use of more GPUs.}
\label{table:utilization}
\scriptsize
\begin{tabular}{|rcccccc|}
\hline
\rowcolor{LightBlue} 
\multicolumn{1}{|c|}{\cellcolor{LightBlue}}                                    & \multicolumn{2}{c|}{\cellcolor{LightBlue}\textbf{Utilization}}                                                  & \multicolumn{2}{c|}{\cellcolor{LightBlue}\textbf{Memory Footprint}}                                                   & \multicolumn{2}{c|}{\cellcolor{LightBlue}\textbf{Bus Utilization}}              \\ \cline{2-7} 
\rowcolor{LightBlue} 
\multicolumn{1}{|c|}{\cellcolor{LightBlue}\textbf{\# GPUs}}& \multicolumn{1}{c|}{\cellcolor{LightBlue}\textbf{CPU}} & \multicolumn{1}{c|}{\cellcolor{LightBlue}\textbf{GPU}} & \multicolumn{1}{c|}{\cellcolor{LightBlue}\textbf{System}} & \multicolumn{1}{c|}{\cellcolor{LightBlue}\textbf{GPU}} & \multicolumn{1}{c|}{\cellcolor{LightBlue}\textbf{PCIe}} & \textbf{NVLink}       \\
\rowcolor{LightBlue} 
              \multicolumn{1}{|c|}{\cellcolor{LightBlue}}
              & \multicolumn{1}{c|}{\cellcolor{LightBlue}(\%)}         & \multicolumn{1}{c|}{\cellcolor{LightBlue}(\%)}         & \multicolumn{1}{c|}{\cellcolor{LightBlue}(MB)}            & \multicolumn{1}{c|}{\cellcolor{LightBlue}(MB)}         & \multicolumn{1}{c|}{\cellcolor{LightBlue}(MBps)}        & (MBps)                \\ \hline
\multicolumn{7}{|c|}{\cellcolor[HTML]{ECF4FF}\textbf{MLPf\_Res50\_TF}} \\\hline
               \multicolumn{1}{|c|}{1xV100}              & \multicolumn{1}{c|}{10.76}        & \multicolumn{1}{c|}{85.84}        & \multicolumn{1}{c|}{17,922}          & \multicolumn{1}{c|}{15,927}       & \multicolumn{1}{c|}{1,251}         & 0                     \\
               \multicolumn{1}{|c|}{2xV100}              & \multicolumn{1}{c|}{16.25}        & \multicolumn{1}{c|}{188.08}       & \multicolumn{1}{c|}{18,521}          & \multicolumn{1}{c|}{31,896}       & \multicolumn{1}{c|}{2,609}         & 967                       \\
               \multicolumn{1}{|c|}{4xV100}              & \multicolumn{1}{c|}{29.06}        & \multicolumn{1}{c|}{372.43}       & \multicolumn{1}{c|}{19,970}          & \multicolumn{1}{c|}{62,214}       & \multicolumn{1}{c|}{4,269}         & 2,867                 \\ \hline
\multicolumn{7}{|c|}{\cellcolor[HTML]{ECF4FF}\textbf{MLPf\_Res50\_MX}} \\\hline
               \multicolumn{1}{|c|}{1xV100}              & \multicolumn{1}{c|}{4.56}         & \multicolumn{1}{c|}{85.84}        & \multicolumn{1}{c|}{7,091}           & \multicolumn{1}{c|}{10,343}       & \multicolumn{1}{c|}{1,251}         & 0                     \\
               \multicolumn{1}{|c|}{2xV100}              & \multicolumn{1}{c|}{9.16}         & \multicolumn{1}{c|}{190.90}       & \multicolumn{1}{c|}{14,924}          & \multicolumn{1}{c|}{20,605}       & \multicolumn{1}{c|}{6,913}         &  1,871                     \\
               \multicolumn{1}{|c|}{4xV100}              & \multicolumn{1}{c|}{18.12}        & \multicolumn{1}{c|}{378.94}       & \multicolumn{1}{c|}{28,781}          & \multicolumn{1}{c|}{40,959}       & \multicolumn{1}{c|}{11,480}        & 21,755                \\ \hline
\multicolumn{7}{|c|}{\cellcolor[HTML]{ECF4FF}\textbf{MLPf\_SSD\_Py}} \\\hline
               \multicolumn{1}{|c|}{1xV100}              & \multicolumn{1}{c|}{3.89}         & \multicolumn{1}{c|}{96.13}        & \multicolumn{1}{c|}{4,100}           & \multicolumn{1}{c|}{15,406}       & \multicolumn{1}{c|}{4,720}         & 0                     \\
               \multicolumn{1}{|c|}{2xV100}              & \multicolumn{1}{c|}{7.21}         & \multicolumn{1}{c|}{180.58}       & \multicolumn{1}{c|}{10,305}          & \multicolumn{1}{c|}{30,772}       & \multicolumn{1}{c|}{6,998}         & 509                      \\
               \multicolumn{1}{|c|}{4xV100}              & \multicolumn{1}{c|}{13.69}        & \multicolumn{1}{c|}{334.84}       & \multicolumn{1}{c|}{20,273}          & \multicolumn{1}{c|}{60,539}       & \multicolumn{1}{c|}{9,791}         & 1,500                 \\ \hline
\multicolumn{7}{|c|}{\cellcolor[HTML]{ECF4FF}\textbf{MLPf\_MRCNN\_Py}} \\\hline
               \multicolumn{1}{|c|}{1xV100}              & \multicolumn{1}{c|}{2.45}         & \multicolumn{1}{c|}{62.46}        & \multicolumn{1}{c|}{7,208}           & \multicolumn{1}{c|}{4,762}        & \multicolumn{1}{c|}{258}           & 0                     \\
               \multicolumn{1}{|c|}{2xV100}              & \multicolumn{1}{c|}{4.83}         & \multicolumn{1}{c|}{144.40}       & \multicolumn{1}{c|}{13,561}          & \multicolumn{1}{c|}{15,933}       & \multicolumn{1}{c|}{2,219}         &  2,472                     \\
               \multicolumn{1}{|c|}{4xV100}              & \multicolumn{1}{c|}{10.39}        & \multicolumn{1}{c|}{283.88}       & \multicolumn{1}{c|}{24,923}          & \multicolumn{1}{c|}{33,935}       & \multicolumn{1}{c|}{3,444}         & 6,547                 \\ \hline
\multicolumn{7}{|c|}{\cellcolor[HTML]{ECF4FF}\textbf{MLPf\_XFMR\_Py}} \\\hline
               \multicolumn{1}{|c|}{1xV100}              & \multicolumn{1}{c|}{1.80}         & \multicolumn{1}{c|}{91.14}        & \multicolumn{1}{c|}{3,992}           & \multicolumn{1}{c|}{14,926}       & \multicolumn{1}{c|}{47}            & 0                     \\
               \multicolumn{1}{|c|}{2xV100}              & \multicolumn{1}{c|}{3.35}         & \multicolumn{1}{c|}{189.30}       & \multicolumn{1}{c|}{7,167}           & \multicolumn{1}{c|}{29,493}       & \multicolumn{1}{c|}{123}           & 11,247                       \\
               \multicolumn{1}{|c|}{4xV100}              & \multicolumn{1}{c|}{6.39}         & \multicolumn{1}{c|}{376.91}       & \multicolumn{1}{c|}{14,244}          & \multicolumn{1}{c|}{58,229}       & \multicolumn{1}{c|}{249}           & 35,862                \\ \hline
\multicolumn{7}{|c|}{\cellcolor[HTML]{ECF4FF}\textbf{MLPf\_GNMT\_Py}} \\\hline
               \multicolumn{1}{|c|}{1xV100}              & \multicolumn{1}{c|}{1.91}         & \multicolumn{1}{c|}{89.94}        & \multicolumn{1}{c|}{7,210}           & \multicolumn{1}{c|}{12,098}       & \multicolumn{1}{c|}{2,743}         & 0                     \\
               \multicolumn{1}{|c|}{2xV100}              & \multicolumn{1}{c|}{3.32}         & \multicolumn{1}{c|}{185.71}       & \multicolumn{1}{c|}{13,561}          & \multicolumn{1}{c|}{24,479}       & \multicolumn{1}{c|}{4,609}         &     1508                  \\
               \multicolumn{1}{|c|}{4xV100}              & \multicolumn{1}{c|}{6.41}         & \multicolumn{1}{c|}{360.89}       & \multicolumn{1}{c|}{24,923}          & \multicolumn{1}{c|}{46,016}       & \multicolumn{1}{c|}{7,692}         & 33,262                \\ \hline
\multicolumn{7}{|c|}{\cellcolor[HTML]{ECF4FF}\textbf{MLPf\_NCF\_Py}} \\\hline
               \multicolumn{1}{|c|}{1xV100}              & \multicolumn{1}{c|}{0.76}         & \multicolumn{1}{c|}{96.39}        & \multicolumn{1}{c|}{1,550}           & \multicolumn{1}{c|}{13,870}       & \multicolumn{1}{c|}{42}            & 0                     \\
               \multicolumn{1}{|c|}{2xV100}              & \multicolumn{1}{c|}{2.41}         & \multicolumn{1}{c|}{194.44}       & \multicolumn{1}{c|}{3,077}           & \multicolumn{1}{c|}{24,847}       & \multicolumn{1}{c|}{110}           &  17,887                     \\
               \multicolumn{1}{|c|}{4xV100}              & \multicolumn{1}{c|}{5.69}         & \multicolumn{1}{c|}{333.11}       & \multicolumn{1}{c|}{5,978}           & \multicolumn{1}{c|}{39,634}       & \multicolumn{1}{c|}{200}           & 75,051                \\ \hline
\multicolumn{7}{|c|}{\cellcolor[HTML]{ECF4FF}\textbf{Dawn\_Res18\_Py}} \\\hline
               \multicolumn{1}{|c|}{1xV100}              & \multicolumn{1}{c|}{4.67}         & \multicolumn{1}{c|}{76.90}        & \multicolumn{1}{c|}{2,670}           & \multicolumn{1}{c|}{2,056}        & \multicolumn{1}{c|}{176}           & 0                     \\ \hline
\multicolumn{7}{|c|}{\cellcolor[HTML]{ECF4FF}\textbf{Dawn\_DrQA\_Py}}  \\\hline
               \multicolumn{1}{|c|}{1xV100}              & \multicolumn{1}{c|}{48.84}        & \multicolumn{1}{c|}{20.30}        & \multicolumn{1}{c|}{6,721}           & \multicolumn{1}{c|}{2,657}        & \multicolumn{1}{c|}{52}            & 0                     \\ \hline
\multicolumn{7}{|c|}{\cellcolor[HTML]{ECF4FF}\textbf{Deep\_GEMM\_Cu}} \\\hline
               \multicolumn{1}{|c|}{1xV100}              & \multicolumn{1}{c|}{1.80}         & \multicolumn{1}{c|}{99.60}        & \multicolumn{1}{c|}{333}             & \multicolumn{1}{c|}{1,067}        & \multicolumn{1}{c|}{13}            & 0                     \\ \hline
\multicolumn{7}{|c|}{\cellcolor[HTML]{ECF4FF}\textbf{Deep\_Conv\_Cu}} \\\hline
               \multicolumn{1}{|c|}{1xV100}              & \multicolumn{1}{c|}{1.73}         & \multicolumn{1}{c|}{99.10}        & \multicolumn{1}{c|}{948}             & \multicolumn{1}{c|}{783}          & \multicolumn{1}{c|}{13}            & 0                     \\ \hline
\multicolumn{7}{|c|}{\cellcolor[HTML]{ECF4FF}\textbf{Deep\_RNN\_Cu}} \\\hline
               \multicolumn{1}{|c|}{1xV100}              & \multicolumn{1}{c|}{1.80}         & \multicolumn{1}{c|}{94.80}        & \multicolumn{1}{c|}{994}             & \multicolumn{1}{c|}{2,536}        & \multicolumn{1}{c|}{3,747}         & 0                     \\ \hline
\multicolumn{7}{|c|}{\cellcolor[HTML]{ECF4FF}\textbf{Deep\_Red\_Cu}} \\\hline
               \multicolumn{1}{|c|}{1xV100}              & \multicolumn{1}{c|}{0.75}         & \multicolumn{1}{c|}{91.30}        & \multicolumn{1}{c|}{313}             & \multicolumn{1}{c|}{631}          & \multicolumn{1}{c|}{27}            & 0                     \\ 
               \multicolumn{1}{|c|}{2xV100}              & \multicolumn{1}{c|}{0.96}         & \multicolumn{1}{c|}{193.20}        & \multicolumn{1}{c|}{430}             & \multicolumn{1}{c|}{994}          & \multicolumn{1}{c|}{86}            & 77,992                     \\ 
               \multicolumn{1}{|c|}{4xV100}              & \multicolumn{1}{c|}{1.68}         & \multicolumn{1}{c|}{366.24}        & \multicolumn{1}{c|}{1123}             & \multicolumn{1}{c|}{2320}          & \multicolumn{1}{c|}{134}            & 404,376                     \\ \hline

\end{tabular}
\vspace{-5mm}
\end{table}

%% file: conclusion.tex
We have presented a detailed characterization of the recent MLPerf benchmark suite in this paper.
{{While MLPerf benchmark characteristics may be heavily influenced by the specific implementations, the suite does provide a diverse set of benchmarks which allows to unveil various bottlenecks in the system. Our experiments point towards (i) the importance of powerful interconnects in  multi-GPU systems,  (ii) the variation in scalability exhibited by different ML models, (iii) the opportunity for smart scheduling strategies in multi-gpu training exploiting the variability in scaling, and (iv) the need for powerful CPUs as host  with increase in number of GPUs.
}}

 We also present the dissimilarity of the benchmarks to other benchmarks in the suite (intra-suite dissimilarity) and dissimilarity against other suites such as DAWNBench and DeepBench (inter-suite dissimilarity). 
 MLPerf provides benchmarks with moderately high memory transactions per second and moderately high compute rates.
 DAWNBench creates a high-compute benchmark with low memory transaction rate, whereas DeepBench provides low compute rate benchmarks.
The various MLPerf benchmarks show uniqueness such as high NVLink utilization in NCF\_Py, low NVLink utilization in SSD\_Py, near-perfect scalability with increasing GPU counts in Res50\_TF and SSD\_Py, and low scalability in NCF\_Py. MRCNN\_Py makes only 1.5$\times$ improvement with tensor cores and reduced precision, whereas Res50\_TF make 3.3$\times$ improvement. 
The DrQA\_Py from DAWNBench results in high CPU utilization and low GPU utilization.

%% file: classification.tex
\begin{landscape}
\begin{table}[p]
\footnotesize
\renewcommand{\arraystretch}{0.8}
\caption{The classification of kernels from 7 benchmarks from MLPerf v0.5.0 submission. Classes are sorted by time.}
\addtocounter{table}{-1}
\centering
\begin{tabular}{R{0.018\textwidth}L{0.29\textwidth}R{0.07\textwidth}R{0.07\textwidth}R{0.10\textwidth}R{0.07\textwidth}R{0.06\textwidth}R{0.15\textwidth}R{0.125\textwidth}R{0.088\textwidth}R{0.088\textwidth}}

  \toprule
 & & \multicolumn{3}{c}{time} & & & & & \multicolumn{1}{c}{intensity} & \multicolumn{1}{c}{throughput} \\
 \cmidrule(lr){3-5}
 \multicolumn{2}{c}{class} & \multicolumn{1}{c}{\%} & \multicolumn{1}{c}{acc\%} & \multicolumn{1}{c}{ms} & \multicolumn{1}{c}{\#call} & \multicolumn{1}{c}{\#unique} & \multicolumn{1}{c}{FLOPs} & \multicolumn{1}{c}{transactions} & \multicolumn{1}{c}{(FLOPs/byte)} & \multicolumn{1}{c}{(GFLOPs/sec)} \\
 \midrule
 & relu & 17.62\% & 17.62\% & 18482.42 & 62,426 & 22 & 8,063,786,730,942 & 198,032,803,079 & 1.27 & 436.29  \\
 \cmidrule(lr){1-2}
\multirow{11}{*}{\rotatebox[origin=c]{90}{MM}} & MM\_256x128 & 6.60\% & 24.22\% & 6924.74 & 22,855 & 6 & 286,017,276,190 & 45,463,681,015 & 0.20 & 41.30  \\
 & MM\_128x128 & 3.94\% & 28.16\% & 4134.53 & 16,360 & 3 & 401,432,676,657 & 37,149,153,847 & 0.34 & 97.09  \\
 & MM\_128x64 & 2.81\% & 30.97\% & 2944.66 & 37,642 & 10 & 11,202,912,520,076 & 19,259,247,078 & 18.18 & 3804.48  \\
 & MM\_256x64 & 0.39\% & 31.35\% & 404.64 & 3,967 & 1 & 30,191,909,672 & 3,727,342,569 & 0.25 & 74.61  \\
 & MM\_64x64 & 0.37\% & 31.73\% & 393.19 & 1,787 & 3 & 18,356,862,434 & 6,165,125,205 & 0.09 & 46.69  \\
\bottomrule
\end{tabular}
\end{table}
\end{landscape}


\begin{landscape}
\begin{table}[p]
\footnotesize
\renewcommand{\arraystretch}{0.8}
\caption{The classification of kernels from 7 benchmarks from MLPerf v0.5.0 submission. Classes are sorted by time. (continue)}
\label{table:classification}
\centering
\begin{tabular}{R{0.018\textwidth}L{0.29\textwidth}R{0.07\textwidth}R{0.07\textwidth}R{0.10\textwidth}R{0.07\textwidth}R{0.06\textwidth}R{0.15\textwidth}R{0.125\textwidth}R{0.088\textwidth}R{0.088\textwidth}}

  \toprule
 & & \multicolumn{3}{c}{time} & & & & & \multicolumn{1}{c}{intensity} & \multicolumn{1}{c}{throughput} \\
 \cmidrule(lr){3-5}
 \multicolumn{2}{c}{class} & \multicolumn{1}{c}{\%} & \multicolumn{1}{c}{acc\%} & \multicolumn{1}{c}{ms} & \multicolumn{1}{c}{\#call} & \multicolumn{1}{c}{\#unique} & \multicolumn{1}{c}{FLOPs} & \multicolumn{1}{c}{transactions} & \multicolumn{1}{c}{(FLOPs/byte)} & \multicolumn{1}{c}{(GFLOPs/sec)} \\
 \midrule
 \multirow{7}{*}{\rotatebox[origin=c]{90}{Pointwise}} & Pointwise\_BitOr & 0.00\% & 72.82\% & 0.62 & 400 & 1 & 0 & 18,800 & 0.00 & 0.00  \\
 & Pointwise\_Clamp & 0.00\% & 72.83\% & 0.60 & 400 & 1 & 0 & 16,400 & 0.00 & 0.00  \\
 & Pointwise\_\_floor\_Float\_ & 0.00\% & 72.83\% & 0.60 & 400 & 1 & 0 & 12,800 & 0.00 & 0.00  \\
 & Pointwise\_MaskedCopy & 0.00\% & 72.83\% & 0.43 & 200 & 1 & 0 & 61,000 & 0.00 & 0.00  \\
 & Pointwise\_GE & 0.00\% & 72.83\% & 0.18 & 100 & 1 & 0 & 23,500 & 0.00 & 0.00  \\
 & Pointwise\_MinValue & 0.00\% & 72.83\% & 0.17 & 100 & 1 & 0 & 15,000 & 0.00 & 0.00  \\
\bottomrule
\end{tabular}
\end{table}

\end{landscape}


\begin{landscape}
\begin{table}[p]
\footnotesize
\renewcommand{\arraystretch}{0.8}
\caption{The comparison between the kernel classifications of the 7 benchmarks from MLPerf v0.5.0 submission. Classes are sorted by time.}
\addtocounter{table}{-1}
\centering
\begin{tabular}{R{0.018\textwidth}L{0.29\textwidth}*{7}{R{0.06\textwidth}@{}R{0.06\textwidth}}}

 \toprule
 & &\multicolumn{2}{c}{Res50\_TF}&\multicolumn{2}{c}{Res50\_MX}&\multicolumn{2}{c}{XFMR\_Py}&\multicolumn{2}{c}{GNMT\_Py}&\multicolumn{2}{c}{SSD\_Py}&\multicolumn{2}{c}{MRCNN\_Py}&\multicolumn{2}{c}{NCF\_Py} \\
 \cmidrule(lr){3-4} \cmidrule(lr){5-6} \cmidrule(lr){7-8} \cmidrule(lr){9-10} \cmidrule(lr){11-12} \cmidrule(lr){13-14} \cmidrule(lr){15-16}
 \multicolumn{2}{c}{class} & time\% & \#call & time\% & \#call & time\% & \#call & time\% & \#call & time\% & \#call & time\% & \#call & time\% & \#call \\
 \midrule
                                                           & relu                                     & 22.90\%&   17,000& 33.06\%&   10,142&  0.00\%&        0&  0.00\%&        0& 25.12\%&    6,100& 13.69\%&   29,184&  0.00\%&        0 \\
\bottomrule
\end{tabular}
\end{table}
\end{landscape}

\begin{landscape}
\begin{table}[p]
\footnotesize
\renewcommand{\arraystretch}{0.8}
\caption{The comparison between the kernel classifications of the 7 benchmarks from MLPerf v0.5.0 submission. Classes are sorted by time.}
\label{table:perbench_classification}
\centering
\begin{tabular}{R{0.018\textwidth}L{0.29\textwidth}*{7}{R{0.06\textwidth}@{}R{0.06\textwidth}}}

 \toprule
 & &\multicolumn{2}{c}{Res50\_TF}&\multicolumn{2}{c}{Res50\_MX}&\multicolumn{2}{c}{XFMR\_Py}&\multicolumn{2}{c}{GNMT\_Py}&\multicolumn{2}{c}{SSD\_Py}&\multicolumn{2}{c}{MRCNN\_Py}&\multicolumn{2}{c}{NCF\_Py} \\
 \cmidrule(lr){3-4} \cmidrule(lr){5-6} \cmidrule(lr){7-8} \cmidrule(lr){9-10} \cmidrule(lr){11-12} \cmidrule(lr){13-14} \cmidrule(lr){15-16}
 \multicolumn{2}{c}{class} & time\% & \#call & time\% & \#call & time\% & \#call & time\% & \#call & time\% & \#call & time\% & \#call & time\% & \#call \\
 \midrule
\multirow{7}{*}{\rotatebox[origin=c]{90}{Pointwise     }} & Pointwise\_BitOr                         &  0.00\%&        0&  0.00\%&        0&  0.00\%&        0&  0.00\%&        0&  0.00\%&        0&  0.00\%&      400&  0.00\%&        0 \\
\bottomrule
\end{tabular}
\end{table}
\end{landscape}

\begin{landscape}
\begin{table}[p]
\footnotesize
\renewcommand{\arraystretch}{0.8}
\caption{Unique kernel sources distribution for each kernel class. Sorted by number of unique kernels. }
\addtocounter{table}{-1}
\centering
\begin{tabular}{R{0.018\textwidth}L{0.29\textwidth}C{0.06\textwidth}*{7}{C{0.06\textwidth}}}

 \toprule
 \multicolumn{2}{c}{class} &\multicolumn{1}{c}{all 7} &\multicolumn{1}{c}{Res50\_TF}&\multicolumn{1}{c}{Res50\_MX}&\multicolumn{1}{c}{XFMR\_Py}&\multicolumn{1}{c}{GNMT\_Py}&\multicolumn{1}{c}{SSD\_Py}&\multicolumn{1}{c}{MRCNN\_Py}&\multicolumn{1}{c}{NCF\_Py} \\
 \midrule
                                                           & fusion                                   & 373 &373&  0&  0&  0&  0&  0&  0 \\
 \cmidrule(lr){1-2}
\multirow{35}{*}{\rotatebox[origin=c]{90}{Pointwise     }} & Pointwise\_Copy                          &  22 &  0&  0&  2&  7& 11& 14&  4 \\
                                                           & Pointwise\_Take                          &   5 &  0&  0&  2&  1&  0&  4&  0 \\
                                                           & Pointwise\_MaxValue                      &   4 &  0&  0&  0&  0&  2&  4&  1 \\
                                                           & Pointwise\_Put                           &   3 &  0&  0&  0&  0&  1&  3&  0 \\
                                                           & Pointwise\_ThresholdUpdateGradInput      &   3 &  0&  0&  1&  0&  2&  1&  2 \\
                                                           & Pointwise\_Fill                          &   3 &  0&  0&  3&  2&  2&  3&  1 \\
\bottomrule
\end{tabular}
\end{table}
\end{landscape}
\begin{landscape}
\begin{table}[p]
\footnotesize
\renewcommand{\arraystretch}{0.8}
\caption{Unique kernel sources distribution for each kernel class. Sorted by number of unique kernels. (continue)}
\label{table:unique_source}
\centering
\begin{tabular}{R{0.018\textwidth}L{0.29\textwidth}C{0.06\textwidth}*{7}{C{0.06\textwidth}}}

 \toprule
 \multicolumn{2}{c}{class} &\multicolumn{1}{c}{all 7} &\multicolumn{1}{c}{Res50\_TF}&\multicolumn{1}{c}{Res50\_MX}&\multicolumn{1}{c}{XFMR\_Py}&\multicolumn{1}{c}{GNMT\_Py}&\multicolumn{1}{c}{SSD\_Py}&\multicolumn{1}{c}{MRCNN\_Py}&\multicolumn{1}{c}{NCF\_Py} \\
 \midrule
\multirow{4}{*}{\rotatebox[origin=c]{90}{cub           }} & cub\_CopyIf                              &   1 &  0&  0&  1&  1&  1&  1&  0 \\
                                                           & cub\_Scan                                &   1 &  0&  0&  1&  0&  0&  1&  0 \\
                                                           & cub\_RadixSortScanBins                   &   1 &  0&  0&  0&  0&  0&  1&  0 \\
                                                           & subtotal of 10                           &  30 &  0&  0& 11& 10& 16& 19&  4 \\
 \cmidrule(lr){1-2}
                                                           & tensorIter                               &  24 &  0&  0& 11& 14& 15& 17&  5 \\
 \cmidrule(lr){1-2}
                                                           & relu                                     &  22 &  5& 10&  0&  0&  9& 10&  0 \\
 \cmidrule(lr){1-2}
                                                           & normalize                                &  13 &  0&  9&  0&  0&  4&  0&  0 \\
\bottomrule
\end{tabular}
\end{table}
\end{landscape}